\documentclass[lettersize,journal]{IEEEtran}
\usepackage{amsmath,amsfonts}
\usepackage{algorithm}
\usepackage{algpseudocode}
\usepackage{array}
\usepackage[caption=false,font=footnotesize]{subfig}
\usepackage{textcomp}
\usepackage{stfloats}
\usepackage{url}
\usepackage{verbatim}
\usepackage{graphicx}
\usepackage{cite}
\usepackage{amsmath}
\usepackage{amssymb}
\usepackage{xcolor}
\usepackage{booktabs}
\usepackage{multirow}
\usepackage{bbding}
\usepackage{orcidlink}
\usepackage{bbm}
\usepackage{colortbl}
\usepackage[capitalise]{cleveref}
\usepackage{graphicx}
\usepackage{epstopdf}
\usepackage{tabularx}
\usepackage{needspace}
\usepackage{placeins}
\AtBeginDocument{%
}
\begin{document}

\title{GLCR: Global--Local Collaborative Routing for Multi-Task Electromagnetic Signal Processing}

\author{Fang Lu\orcidlink{0009-0003-4614-3533}, Yao Lu\orcidlink{0000-0003-0655-7814}, ~\IEEEmembership{Student Member, IEEE}, Zhuangzhi Chen\orcidlink{0000-0002-8736-8814}, Sicheng Zhang\orcidlink{0000-0002-9088-8773},~\IEEEmembership{Member,~IEEE}, Shanqing Yu\orcidlink{0000-0001-5170-8082}, Yun Lin\orcidlink{0000-0003-1379-9301},~\IEEEmembership{Senior Member, IEEE}, Qi Xuan\orcidlink{0000-0002-6320-7012},~\IEEEmembership{Senior Member,~IEEE}, Guan Gui\orcidlink{0000-0003-3888-2881},~\IEEEmembership{Fellow,~IEEE}
\thanks{This work was partially supported by the National Natural Science Foundation of China under Grant (62301492,62503423), the Baima Lake Laboratory Joint Fund of the Zhejiang Provincial Natural Science Foundation of China (LBMHZ25F020002), the National Key R\&D Program of China (2025YFA1510900), the Zhejiang Provincial Key R\&D Program (2024C01025). (Corresponding author: Yao Lu)}
\thanks{Fang Lu, Yao Lu, Shanqing Yu and Qi Xuan are with the Institute of Cyberspace Security, College of Information Engineering, Zhejiang University of Technology, Hangzhou 310023, China, also with the Binjiang Institute of Artificial Intelligence, Zhejiang University of Technology, Hangzhou 310056, China.}
\thanks{Zhuangzhi Chen is with the Institute of Cyberspace Security, Zhejiang University of Technology, Hangzhou 310023, China, and also with the Research Center of Electromagnetic Space Security, Binjiang Institute of Artificial Intelligence, ZJUT, Hangzhou, 310056, China.}
\thanks{Sicheng Zhang and Yun Lin are with the College of Information and Communication Engineering, Harbin Engineering University, Harbin 150001, China.}
\thanks{Guan Gui is with the College of Telecommunications and Information Engineering, Nanjing University of Posts and Telecommunications, Nanjing 210003, China.}}

\markboth{IEEE Transactions on Communications}%
{Lu \MakeLowercase{\textit{et al.}}: GLCR for Multi-Task Electromagnetic Signal Processing}

\maketitle
\begin{abstract}
Deep neural networks for electromagnetic signal processing are typically developed
as single-task experts with fixed inputs and objectives, exhibiting limited
cross-task transferability. In contrast, electromagnetic foundation models offer
broader task coverage but incur substantial pretraining and inference costs.
To balance accuracy and efficiency in multi-task processing, we adopt a model
routing mechanism that dynamically dispatches unknown signals to matched experts
within a heterogeneous pool. To address the lack of public benchmarks and
systematic studies for this mechanism, we construct HeteroSignal-7, a multi-task
routing benchmark with 140,000 samples from seven datasets covering six task types,
paired with seven classification and waveform-recovery experts and a unified
evaluation protocol.
We then propose GLCR, a two-stage routing method that combines efficient global
dispatch with reliable local discrimination of similar tasks.
In the first stage, a lightweight one-dimensional convolutional
neural network (CNN) with an additive-margin head predicts one
of seven routing classes. In the second stage, a gradient-boosted-tree
specialist uses four physical statistics to re-evaluate samples
whose top two candidates match a confusion pair automatically
identified from validation errors. The pipeline is trained
across the full signal-to-noise ratio (SNR) range and requires
neither SNR labels nor a noise estimation module at inference time.
GLCR outperforms all 19 baselines, achieving 96.16\% routing accuracy
and a 76.30\% system score, exceeding the respective second-best
baselines by 1.16\% and 1.14\%. It also achieves
the lowest routing latency among the compared methods, at
1.43~ms per sample, and reduces expected overall system FLOPs
by a factor of 5.9 compared with executing all experts.
This work provides a public evaluation platform and a strong,
efficient, and reproducible baseline for heterogeneous expert
routing in electromagnetic signal processing.

\end{abstract}

\enlargethispage{\baselineskip}

\begin{figure}[!t]
\centering
\includegraphics[width=\columnwidth]{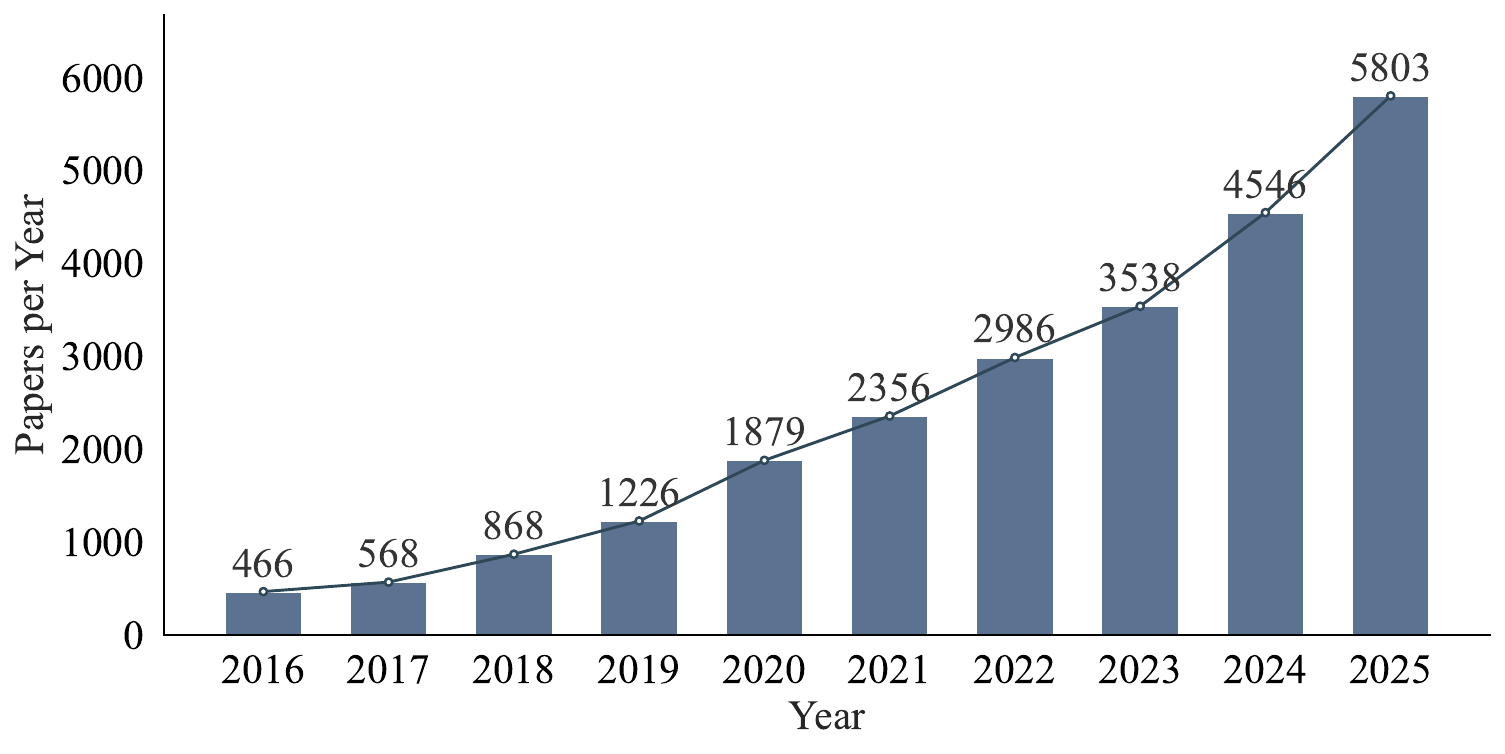}
\caption{Number of publications on machine-learning methods for
communications from 2016 to 2025, retrieved from
OpenAlex~\cite{priem2022openalex}.}
\label{fig:growth}
\end{figure}

\Needspace{6\baselineskip}
\section{Introduction}

\IEEEPARstart{W}{ireless} communication and radar systems involve diverse electromagnetic
signal-processing tasks, ranging from modulation classification and
interference recognition to emitter identification and waveform recovery.
In recent years, deep neural networks (DNNs) have become a mainstream
approach to these tasks, as evidenced by the steady growth in publications
on machine learning for communications from 2016 to 2025 indexed by
OpenAlex~\cite{priem2022openalex} (\cref{fig:growth}). By learning
representations directly from raw electromagnetic signals, DNNs have
achieved strong performance across various individual recognition and
waveform-recovery tasks
~\cite{o2018over,zhang2023jdmr,li2026open,xu2025p3mc}. However, most
existing networks are developed independently for a single task, even as
modern wireless systems increasingly integrate communication and sensing
functions~\cite{liu2022integrated}. These tasks differ in modulation schemes,
channel impairments, and output modalities, ranging from discrete class
labels to continuous waveforms. Consequently, a model optimized for one
task cannot be directly reused for another, hindering effective multi-task
processing.

\begin{figure*}[!t]
\centering
\includegraphics[width=\textwidth]{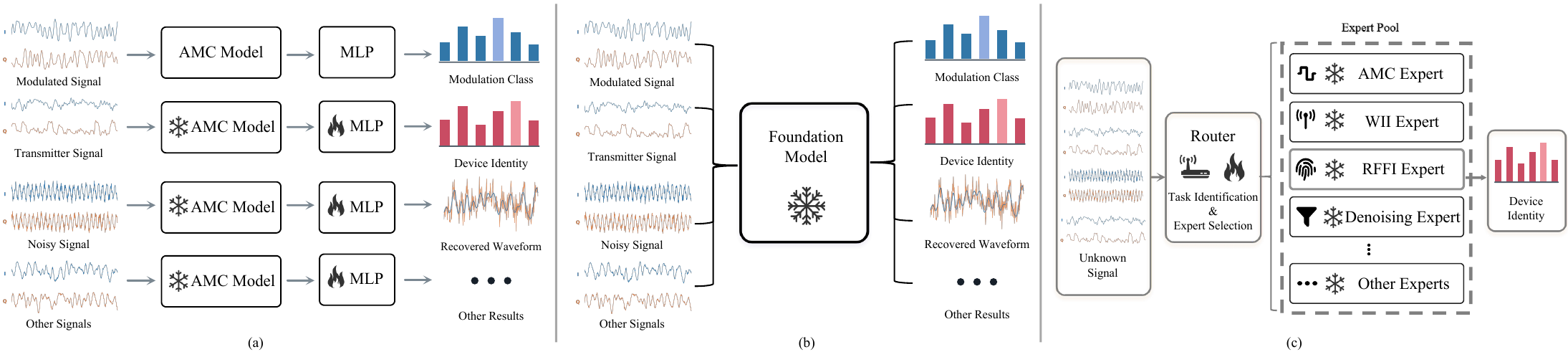}
\caption{Three paradigms for multi-task electromagnetic signal processing:
(a) transfer learning with a frozen source-task backbone and task-specific
heads, (b) an electromagnetic foundation model pretrained on multi-task data
and adapted through task-specific heads, and (c) model routing that selects
task-specific experts from a heterogeneous pool.}
\label{fig:paradigms}
\end{figure*}

\begin{figure*}[!t]
\centering
\subfloat[]{\includegraphics[height=0.19\textheight]{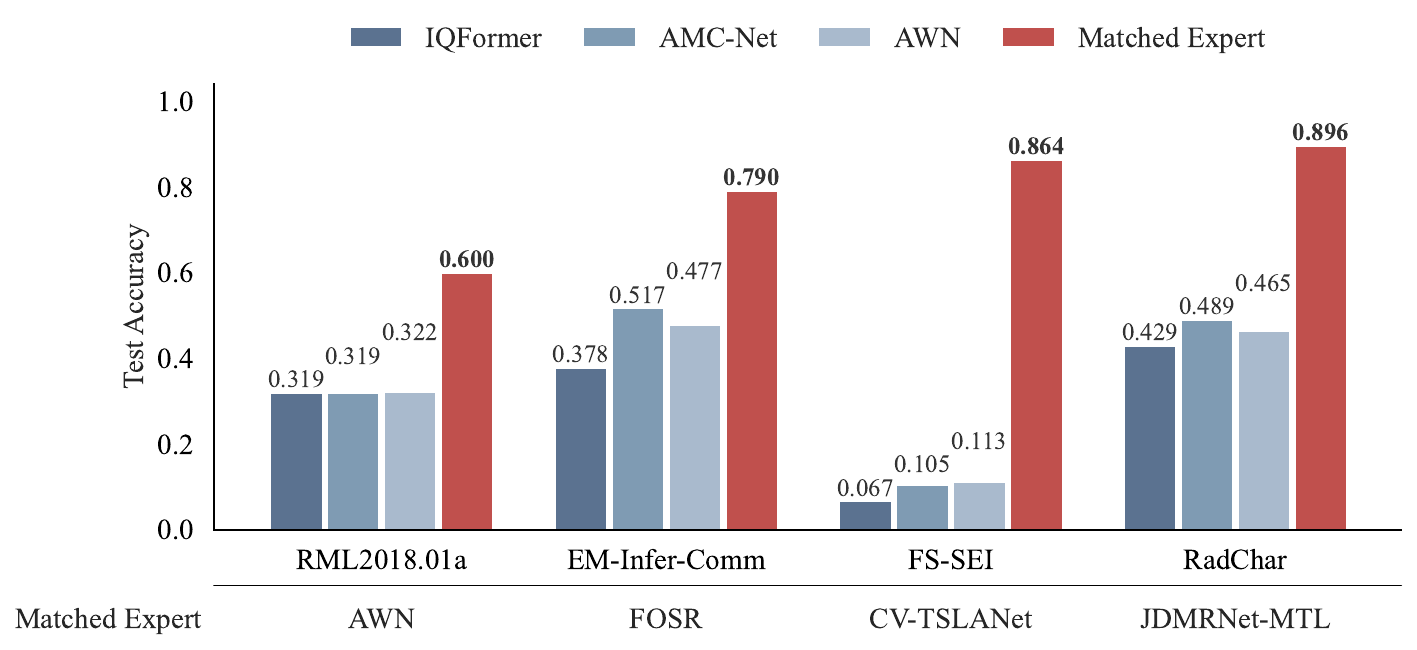}%
\label{fig:headswap}}
\hfill
\subfloat[]{\includegraphics[height=0.19\textheight]{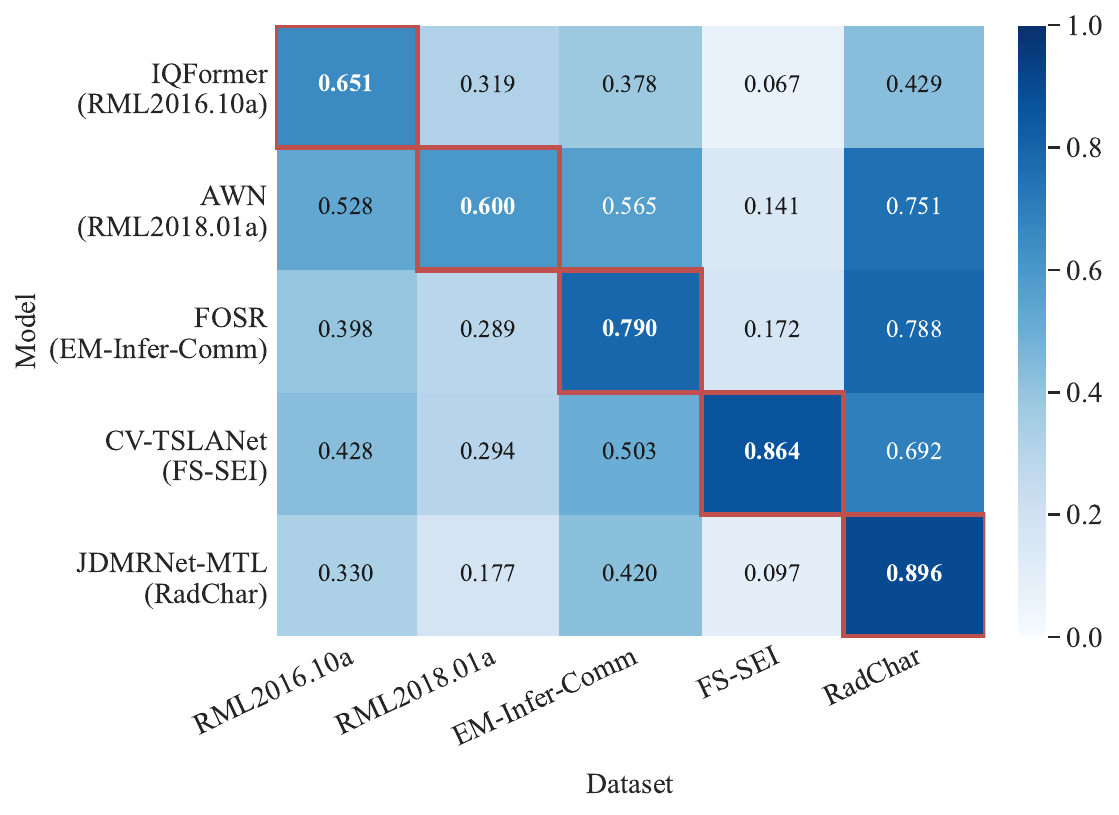}%
\label{fig:cross_matrix}}
\caption{Cross-task transfer with frozen backbones and retrained probe
heads. (a) Cross-dataset transfer of three AMC models trained on
RML2016.10a. (b) Head-swapped cross-evaluation of five classification
experts. Red marks indicate dataset-matched entries, which differ from
the full-expert results in \cref{tab:experts}. Waveform-output
tasks are excluded from (b).}
\label{fig:motivation}
\end{figure*}

To address this multi-task requirement, existing studies mainly explore two
paradigms. The first paradigm adapts source-task models to new tasks through
transfer learning, as illustrated in
\cref{fig:paradigms}(a)~\cite{wong2022transfer,kuzdeba2021transfer}.
A backbone pretrained on a source task is reused with task-specific heads,
with all or part of its parameters frozen. However, differences in modulation
schemes, hardware impairments, and channel conditions can cause feature
mismatch and limit cross-task generalization
~\cite{wong2022transfer,kuzdeba2021transfer}. Our experiments illustrate
this limitation: when a backbone is transferred across datasets,
or when a classification head is paired with a backbone trained for another
task, performance drops relative to the dataset-matched model. In both
experiments, each dataset achieves its best result with its matched
backbone--head combination
(\cref{fig:motivation}(a) and \cref{fig:motivation}(b)). The second paradigm
develops electromagnetic foundation models, as shown in
\cref{fig:paradigms}(b)~\cite{luo2025emind,zhou2025spectrumfm}. These models
pretrain a shared backbone on large-scale multi-task data and adapt it to
downstream tasks through task-specific heads, offering broader cross-task
generalization. However, both large-scale pretraining and shared-backbone
inference can incur substantial computational costs.

To balance multi-task performance and computational efficiency, we explore a
new paradigm, namely model routing over a heterogeneous expert pool, as
illustrated in \cref{fig:paradigms}(c). In this paradigm, a lightweight router
identifies the task of an unknown input signal and dispatches it to the
corresponding frozen expert, avoiding the joint retraining of all models and
executing only the router and the selected expert at inference time. However,
training and fairly evaluating such routers remain difficult because, to our
knowledge, no public benchmark provides heterogeneous tasks, matched experts,
and a unified system-level evaluation protocol for this setting. To bridge
this gap, we first construct HeteroSignal-7, a multi-task electromagnetic
signal routing benchmark that integrates six task types and seven
representative datasets, paired with a heterogeneous pool of seven experts
covering both classification and waveform recovery.

Building upon this benchmark, we further propose Global--Local Collaborative
Routing (GLCR; \cref{sec:method}), a two-stage routing method designed to
achieve both fast global dispatching and reliable discrimination of similar
tasks. Specifically, the first stage employs a lightweight one-dimensional
convolutional neural network (CNN) with an additive-margin classification
head to perform global classification across all seven routing classes.
However, under noise and channel distortion, signals from certain tasks can
exhibit highly similar degraded waveforms, making single-stage global routing
prone to concentrated errors on these confusable tasks. To address this issue,
GLCR identifies confusing task pairs from validation errors. When the top
two candidates match an identified pair, a gradient-boosted tree specialist
uses four physical statistics for local re-evaluation. The entire pipeline
is trained across the full signal-to-noise ratio (SNR) range without
requiring SNR labels or noise estimation at inference time.

The main contributions of this paper are as follows:
\begin{itemize}
\item[$\bullet$] We build HeteroSignal-7, a public evaluation benchmark for
multi-task electromagnetic signal routing. It covers six task types and seven
datasets, and provides a heterogeneous pool of classification and waveform
recovery experts together with a unified evaluation protocol.
\item[$\bullet$] We propose GLCR, a two-stage routing method that uses
lightweight global decisions to dispatch most samples and introduces
local re-evaluation based on physical statistics for confusing task pairs.
It improves discrimination of similar signals while keeping routing costs low.
\item[$\bullet$] We conduct extensive comparisons and ablation experiments
using the unified benchmark and expert pool. The results show that our method
outperforms all evaluated baselines in both routing accuracy and system score, while
reducing the inference cost of the multi-task system.
\end{itemize}

\section{Related Work}
\label{sec:Related Works}

\textbf{Expert Models for Electromagnetic Signal Tasks.}
Early automatic modulation classification (AMC) methods were mainly based on statistical
signal processing theory, including likelihood-based methods~\cite{wei2000maximum} and
feature-based methods~\cite{wu2008novel}.
As computing power and available data have grown, deep learning has become a widely used
approach to AMC. O'Shea et al.~\cite{o2016convolutional}
pioneered the use of convolutional neural networks for modulation recognition from
time-domain in-phase/quadrature (I/Q) signals.
Subsequent studies introduced residual connections and other architectures and evaluated
them on over-the-air signals~\cite{o2018over}.
To capture the temporal structure of baseband signals, recurrent networks such as
LSTM~\cite{hochreiter1997long} and GRU~\cite{hong2017automatic},
and their variants~\cite{ansari2025novel}, were introduced to model long-range
dependencies. Hybrid architectures combine convolutional and recurrent units
~\cite{cheng2024automatic,elsagheer2023hybrid}, or introduce global feature
extraction modules such as attention~\cite{qu2024enhancing}.
Recent work has further focused on feature representation and computational efficiency.
Representative architectures include SigNet~\cite{chen2021signet},
which uses a learnable signal-to-matrix operator; the multi-scale feature extraction
network AMC-Net~\cite{zhang2023amc}; the multimodal fusion model
IQFormer~\cite{shao2024iqformer}; the adaptive wavelet network
AWN~\cite{zhang2023toward}; and the parameter-efficient
PET-CGDNN~\cite{zhang2021efficient}. Beyond compact architecture design,
model pruning offers another route to efficiency, including generic layer
pruning~\cite{lu2024generic} and a two-stage recoverable pruning framework for AMC
models~\cite{lu2026fcos}.

Beyond AMC, specialized models address other electromagnetic signal processing tasks.
Examples include FOSR~\cite{li2026open} for open-set
wireless interference recognition, P3MC~\cite{xu2025p3mc} for few-shot
radio-frequency fingerprint identification, and JDMR-Net~\cite{zhang2023jdmr}
for joint radar detection and modulation recognition. For waveform recovery
tasks such as signal denoising (SD) and blind source separation (BSS),
time-domain end-to-end speech separation architectures such as
Conv-TasNet~\cite{luo2019conv} and DPRNN~\cite{luo2020dual} can also be
adapted to process two-channel I/Q baseband signals. These models are typically
optimized for a single task and differ in native input length, feature extraction
operations, and output type. These differences can hinder cross-task reuse and
transfer~\cite{wong2022transfer,kuzdeba2021transfer}. Rather than redesigning
these experts, we treat them as a frozen expert pool and study a lightweight routing
mechanism for expert selection.

\textbf{Electromagnetic Foundation Models.}
To support more tasks than individual experts can handle, recent studies have used
large-scale pretraining to learn transferable representations of electromagnetic
signals. SpectrumFM~\cite{zhou2025spectrumfm} combines convolution
and multi-head self-attention, uses masked reconstruction and next-slot signal
prediction for self-supervised pretraining, and applies parameter-efficient
fine-tuning to tasks such as modulation classification, wireless technology
classification, spectrum sensing, and anomaly detection.
EMind~\cite{luo2025emind} builds a unified pretraining framework for multi-source,
variable-length I/Q data. It improves training efficiency through
length-adaptive multi-signal packing and dataset weighting, and adapts to
classification, parameter estimation, and waveform recovery through task heads.
These studies demonstrate the potential of pretraining and fine-tuning for
electromagnetic signal processing, but aim to cover multiple tasks with a shared
backbone. In contrast, we retain heterogeneous specialists and route each input to the
required expert.

\textbf{Model Routing.}
Reducing inference overhead within a single model can be achieved through conditional
computation, as in mixture-of-experts (MoE) models~\cite{shazeer2017outrageously},
or through structural compression, such as layer pruning~\cite{lu2026reassessing}.
Beyond single-model optimization, the growing number of pretrained models has motivated
system-level routing among independent models, particularly for large language models
(LLMs).
FrugalGPT~\cite{chen2023frugalgpt} and RouteLLM~\cite{ong2025routellm}
assign queries to models with different capabilities through cascaded calls
or preference-based routers, reducing inference cost. For pools with multiple
candidate models, EmbedLLM~\cite{zhuang2025embedllm} learns compact vector
representations to describe the strengths of candidate models.
GraphRouter~\cite{feng2410graphrouter} builds a heterogeneous graph of tasks,
input queries, and candidate models, and uses a graph neural network to
model their matching relations. CSCR~\cite{shirkavand2026cost} introduces
cost-aware contrastive learning to align query embeddings with expert behavior
fingerprints, balancing performance and cost. UniRoute~\cite{jitkrittum2026universal}
further handles dynamic model pools containing models unseen at test time.
It derives model features from predictions on representative queries, enabling routing
to new models. RouterArena~\cite{lu2026routerarena} provides an open,
unified evaluation platform for systematic comparisons of router performance,
latency, and selection behavior.

Although these routing methods have demonstrated promising performance--cost
trade-offs, they primarily target natural-language queries and typically rely on
text-based semantic representations for model selection. Unlike text,
raw I/Q signals are affected by modulation mechanisms, hardware impairments,
and channel noise. At low SNR, signals from different tasks may have similar time-domain
waveforms, causing errors in single-stage global routing to concentrate on certain task
pairs. To address this issue, GLCR combines global decisions with local
re-evaluation using physical statistics of complex baseband signals.

\section{Problem Formulation}
\label{sec:problem}

Consider a complex baseband signal $s(t)=I(t)+\mathrm j\,Q(t)$ acquired by a receiver.
The signal is sampled at rate $f_s$ and, when necessary, center-cropped or zero-padded
to a fixed-length segment. It is represented by a two-channel real-valued tensor
$\mathbf x\in\mathbb R^{2\times L}$, whose channels are the in-phase (I) and
quadrature (Q) components, and $L$ is the number of sampling points.

Consider $K$ electromagnetic signal processing tasks, each associated with a pretrained,
frozen expert $e_k$. Without prior SNR information, the router $R$ maps an input sample
$\mathbf x$ in the unified format to a predicted task label
$\hat y\in\{1,\dots,K\}$, and sends it to the corresponding expert $e_{\hat y}$.
Only the selected expert is executed at inference time. The routing goal is to maximize
the expected per-sample task utility:
\begin{equation}
R^{*}=\arg\max_{R}\;
\mathbb E_{\mathbf x}\Bigl[\,u\bigl(\mathbf x,e_{R(\mathbf x)}\bigr)\Bigr],
\label{eq:objective}
\end{equation}
where $u(\mathbf x,e)\in[0,1]$ is the utility of processing sample
$\mathbf x$ with expert $e$. For a correctly routed sample, classification utility is
based on prediction correctness, while waveform-recovery utility is based on normalized
improvement. An incorrectly routed sample receives zero utility.

\section{The HeteroSignal-7 Benchmark}
\label{sec:dataset}

\subsection{Datasets and Construction}
\label{sec:dataset:construction}

\begin{figure*}[!t]
\centering
\includegraphics[width=\textwidth]{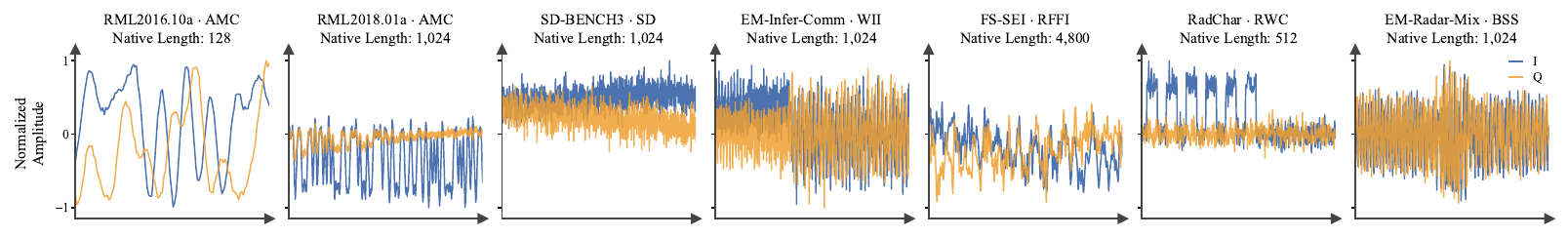}
\caption{Example samples of the seven datasets in the unified
1{,}024-point router input format (amplitude-normalized I/Q).
Zero-padded datasets (RML2016.10a, RadChar) are shown over their
non-zero segments; native lengths are annotated in the panel titles,
and FS-SEI is center-cropped from its native 4{,}800-point sequence.}
\label{fig:gallery}
\end{figure*}

\begin{table*}[!t]
\centering
\caption{The HeteroSignal-7 Benchmark. Bold entries are constructed
in this work using source data or generation protocols from the cited references.}
\label{tab:dataset}
\footnotesize
\renewcommand{\arraystretch}{1.25}
\begin{tabularx}{\textwidth}{@{}llXcccc@{}}
\toprule
Dataset & Task & \multicolumn{1}{c}{Signal Type} & \begin{tabular}[c]{@{}c@{}}Signal\\ Length\end{tabular} & \begin{tabular}[c]{@{}c@{}}Sampling\\ Rate\end{tabular} & \begin{tabular}[c]{@{}c@{}}Data\\ Source\end{tabular} & \begin{tabular}[c]{@{}c@{}}Number of\\ Samples\end{tabular} \\
\midrule
RML2016.10a~\cite{o2016radio} & AMC & 11 modulation classes, SNR: $-20$--$18$~dB & 128 & 1~MHz & Simulation & 20,000 \\
RML2018.01a~\cite{o2018over} & AMC & 24 modulation classes, SNR: $-20$--$30$~dB & 1,024 & 1~MHz & Simulation & 20,000 \\
\textbf{SD-BENCH3}~\cite{o2018over} & SD & 24 modulation classes with AWGN, SNR: $-3$--$20$~dB & 1,024 & 1~MHz & \textbf{Simulation} & 20,000 \\
EM-Infer-Comm~\cite{luo2025emind} & WII & 9 interference types, SNR: $-20$--$20$~dB & 1,024 & 20~MHz & Simulation & 20,000 \\
FS-SEI~\cite{wang2022few} & RFFI & 90 ADS-B emitters, SNR: not annotated & 4,800 & 50~MHz & Real-world & 20,000 \\
RadChar~\cite{huang2023multi} & RWC & 5 radar waveform classes, SNR: $-20$--$20$~dB & 512 & 3.2~MHz & Simulation & 20,000 \\
\textbf{EM-Radar-Mix}~\cite{luo2025emind} & BSS & 8 radar waveform classes, $K{=}2$ mixtures, SNR: $12$~dB & 1,024 & 5~MHz & \textbf{Simulation} & 20,000 \\
\midrule
TOTAL & \multicolumn{5}{c}{7 datasets across 6 task types, unified train/validation/test split of 8:1:1} & \textbf{140,000} \\
\bottomrule
\end{tabularx}
\end{table*}
To the best of our knowledge, no public electromagnetic signal benchmark is designed
specifically to evaluate heterogeneous expert routing (\cref{sec:Related Works}). To
fill this gap, we construct HeteroSignal-7. The benchmark covers six task types across
seven datasets: RML2016.10a and RML2018.01a for automatic modulation classification
(AMC), SD-BENCH3 for signal denoising (SD), EM-Infer-Comm for wireless interference
identification (WII), FS-SEI for radio-frequency fingerprint identification (RFFI),
RadChar for radar waveform classification (RWC), and EM-Radar-Mix for blind source
separation (BSS). Each dataset contains $20{,}000$ samples, yielding $140{,}000$
samples in total. Five datasets are drawn from public sources, and two are constructed
in this work. The benchmark is summarized in \cref{tab:dataset}, and \cref{fig:gallery}
shows one sample from each dataset in the unified routing format. We use $20{,}000$ samples per dataset to balance task priors and provide similar
statistical precision across tasks. Under the $8{:}1{:}1$ split, each dataset initially
contributes $2{,}000$ test samples; for accuracy above $90\%$, the corresponding $95\%$
confidence interval is approximately $\pm1.3\%$ or narrower, balancing statistical
resolution and computational cost.

Among the public sources, RML2016.10a~\cite{o2016radio} and RML2018.01a~\cite{o2018over}
are simulated modulation datasets generated with GNU Radio, containing $11$ and $24$
modulation classes,
respectively. EM-Infer-Comm~\cite{luo2025emind} provides nine types of
wireless interference signals. FS-SEI~\cite{wang2022few} contains ADS-B signals from $90$ real civil aircraft transmitters and provides the only real-world measurements in the benchmark.
RadChar~\cite{huang2023multi} contains five radar waveform classes.

SD-BENCH3 is a denoising benchmark constructed in this work using the
SNR${=}30$~dB samples of RML2018.01a as clean references. It supports controlled
impairment configurations involving I/Q imbalance, carrier frequency offset, and
additive white Gaussian noise (AWGN). For the SD task in HeteroSignal-7, we use the
AWGN-only configuration. Clean references are drawn from the SD-BENCH3 test pool,
which is disjoint from the SD expert's training reference pool, and noisy samples at
multiple SNR levels are generated using independent random seeds. Because the original
EM-Radar-Mix data are not publicly available, we reproduce the waveform types and
mixing protocol of EMind~\cite{luo2025emind} using fixed random seeds. Eight radar
waveform classes are mixed in pairs in the time domain, and noise is added at a sampling
rate of $5$~MHz to give an overall SNR of $12$~dB. The full impairment settings,
waveform parameter ranges, and generation seeds for both datasets are given in
Section~A of the Supplementary Material.

We simulate deployment with a shared acquisition front end.
All received signals pass through the same acquisition pipeline and are delivered to the
router in a unified format: center cropping or
zero-padding to $1{,}024$ points, followed by peak normalization.
Native sampling rates and sequence lengths are dataset-construction artifacts and are
not provided to the router. The router therefore relies solely on signal content in the
unified format, with zero-padding used only to align sequence lengths for batch
processing.

\subsection{Split Protocol and Homology Audit}
\label{sec:dataset:split}

For each dataset, $20{,}000$ samples are selected using stratification by class and SNR;
FS-SEI is stratified only by class because it has no
SNR labels. The samples are then split into training, validation,
and test sets at a fixed $8{:}1{:}1$ ratio using a fixed random seed.
SD samples are grouped by clean reference so that samples derived from the same
reference remain in the same split. Sample-by-sample content
matching also confirms that this reference pool does not overlap
with the SD expert training pool.

Because the clean SD references directly use the original $30$~dB
RML2018.01a signals, we identify exact duplicates across datasets
by SHA-1 content matching after peak normalization, and perform
correlation checks as a supporting audit. Based on this audit, $29$ and $25$ samples
with identified cross-task source overlap are removed from the validation and test sets,
leaving $13{,}971$ and $13{,}975$ samples,
respectively. Checkpoints of all trained models are selected on the
cleaned validation set (Supplementary Material, Section~B). All reported performance
metrics use the same cleaned test set. The audit methods, results, and detection scope
are detailed in Section~C of the Supplementary Material.

Unless otherwise stated, all metrics in this paper are averaged over
all samples. The cleaned test set contains $1{,}990$ samples from
RML2018.01a, $1{,}985$ from SD-BENCH3, and $2{,}000$
from each of the other five datasets.

\begin{table*}[!t]
\centering
\caption{Expert Model Pool of HeteroSignal-7. All experts are trained from scratch.}
\label{tab:experts}
\footnotesize
\renewcommand{\arraystretch}{1.25}
\begin{tabularx}{\textwidth}{@{}llXccccc@{}}
\toprule
Task & Dataset & \multicolumn{1}{c}{Expert Model} & Venue & \begin{tabular}[c]{@{}c@{}}Params\\ (M)\end{tabular} & \begin{tabular}[c]{@{}c@{}}FLOPs\\ (M)\end{tabular} & Test Metric & \begin{tabular}[c]{@{}c@{}}Test\\ Score\end{tabular} \\
\midrule
\multirow{2}{*}{AMC}
    & RML2016.10a & IQFormer~\cite{shao2024iqformer} & IEEE TCCN 2025 & 0.36 & 42 & OA & 65.15\% \\
    & RML2018.01a & AWN~\cite{zhang2023toward} & IEEE TCCN 2023 & 0.38 & 70 & OA & 61.96\% \\
SD  & SD-BENCH3 & Conv-TasNet~\cite{luo2019conv} & IEEE/ACM TASLP 2019 & 2.18 & 293 & SNR gain & $+14.53$~dB \\
WII & EM-Infer-Comm & FOSR~\cite{li2026open} & IEEE TCOM 2026 & 0.82 & 89 & OA & 79.35\% \\
RFFI & FS-SEI & CV-TSLANet~\cite{xu2025p3mc} & IEEE IoT-J 2025 & 6.46 & 889 & OA & 86.75\% \\
RWC & RadChar & JDMRNet-MTL~\cite{zhang2023jdmr} & IEEE TAES 2023 & 1.92 & 54 & OA & 89.45\% \\
BSS & EM-Radar-Mix & DPRNN~\cite{luo2020dual} & ICASSP 2020 & 2.91 & 194 & SI-SNRi & $+23.73$~dB \\
\bottomrule
\end{tabularx}
\end{table*}

\subsection{Expert Model Pool}
\label{sec:experts}

HeteroSignal-7 specifies one fixed expert model for each of its seven
datasets, forming the heterogeneous expert pool shown in
\cref{tab:experts}. Rather than fine-tuning a shared backbone, we select representative,
peer-reviewed models suited to each signal type, using official open-source
implementations or faithful reproductions. All experts are trained from scratch without
external pretrained weights and process their tasks at the native signal length. Expert
evaluation and routing evaluation use the same cleaned test set (\cref{sec:dataset:split}).
Expert parameters remain frozen; the router only selects which expert to execute.

The five classification experts cover modulation, interference,
transmitter, and waveform recognition, with overall accuracy (OA)
as their metric. IQFormer~\cite{shao2024iqformer} and
AWN~\cite{zhang2023toward} serve RML2016.10a and RML2018.01a,
respectively, using their official architectures without modification.
FOSR~\cite{li2026open} was originally designed for open-set recognition. We adapt it for
closed-set classification and
extend its input length from $128$ to $1{,}024$.
For FS-SEI, we use CV-TSLANet, the official backbone of
P3MC~\cite{xu2025p3mc}, attach a linear classification head,
and train the complete model from scratch on the $4{,}800$-point
FS-SEI samples. For JDMRNet-MTL~\cite{zhang2023jdmr}, we replace
the original per-pulse detection branch with four radar parameter
regression heads and reimplement the model in PyTorch to jointly learn waveform
classification and parameter estimation.

The remaining two experts perform signal recovery. The SD expert
Conv-TasNet~\cite{luo2019conv} is taken from the Asteroid codebase
and adapted for single-source enhancement of two-channel I/Q signals.
We reduce its temporal convolution network from the original $24$ blocks to $2\times5$
blocks to handle $1{,}024$-point samples.
Its metric is SNR gain relative to the noisy input. Training samples
are generated from an independent pool of $78{,}624$ clean reference
samples, which has no sample-level content overlap with the SD
reference pool used by the routing benchmark
(Section~C of the Supplementary Material). The BSS expert DPRNN~\cite{luo2020dual}
retains the architecture of the original model, uses two-channel I/Q inputs, and is
trained for two-source separation. Its metric is scale-invariant SNR improvement
(SI-SNRi).

\newcommand{\methodname}{GLCR}

\section{Method}
\label{sec:method}

\begin{figure*}[!t]
\centering
\includegraphics[width=0.9\textwidth]{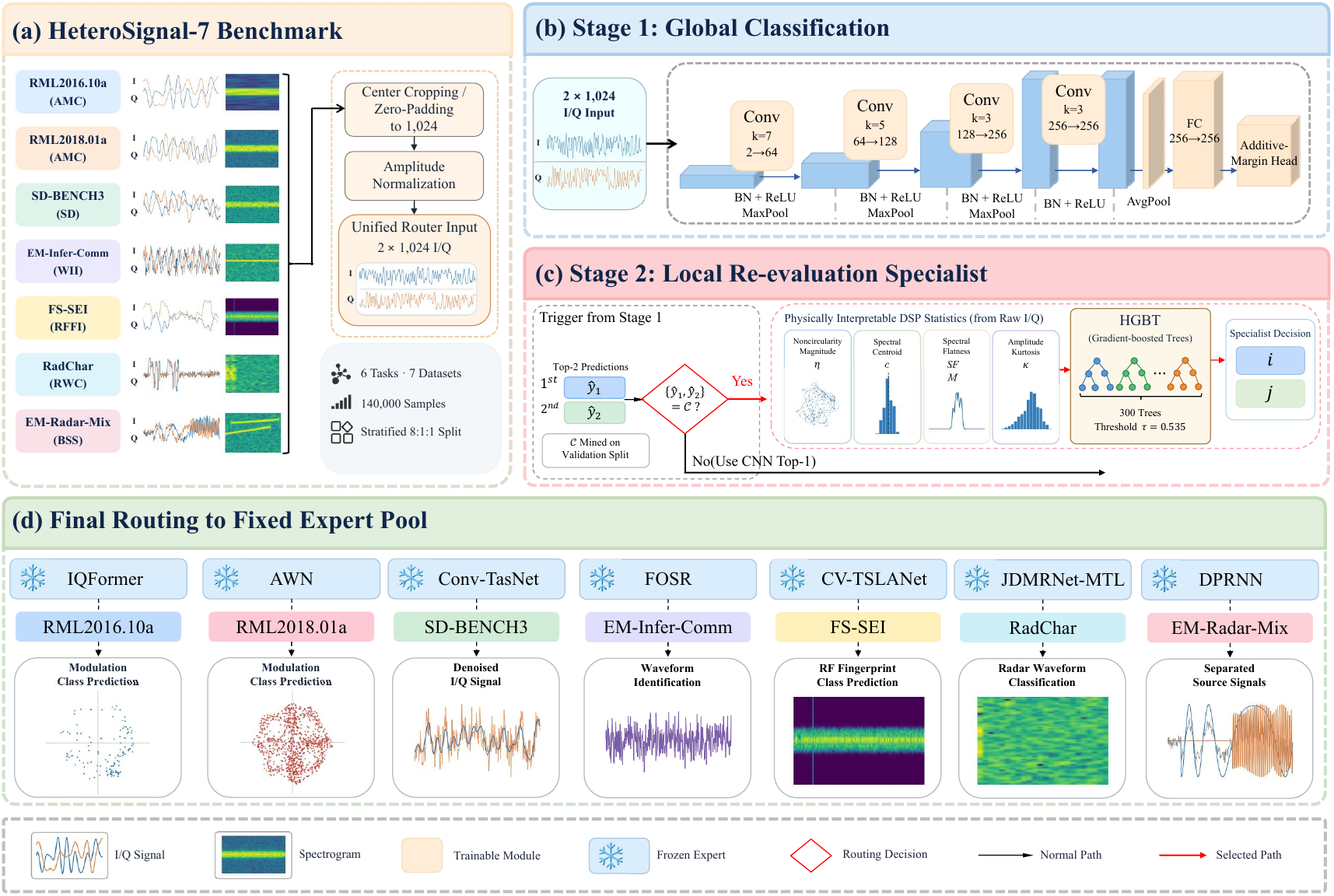}
\caption{Overview of the proposed GLCR framework: (a) unified preprocessing of HeteroSignal-7 samples; (b) Stage-1 global decision using a lightweight CNN with an additive-margin head; (c) top-2 confusion-pair triggering and Stage-2 local re-evaluation using four physical statistics (DSP4) and a binary HGBT specialist; and (d) dispatching to the selected frozen expert.}
\label{fig:framework}
\end{figure*}

Given the HeteroSignal-7 benchmark (\cref{sec:dataset}) and its expert pool, we propose
Global--Local Collaborative Routing (GLCR) to address the routing objective.
The overall framework is shown in \cref{fig:framework}.

Samples are processed independently with frozen experts; \cref{alg:cfr} summarizes the
complete training and inference procedure.

\begin{algorithm}[!t]
\caption{GLCR: Global--Local Collaborative Routing.}
\label{alg:cfr}
\begin{algorithmic}[1]
\Statex \textbf{\# Training}
\State Train the margin-head primary classifier $R$ on the training split
with \cref{eq:margin_head}; freeze $R$.
\State Tally the validation-split routing errors of $R$ by unordered
task pair; take any pair $\mathcal C$ accounting for more than $\rho=0.5$
of all errors as the confusion pair
($\{\mathrm{RML2018.01a},\ \mathrm{SD\mbox{-}BENCH3}\}$ in this benchmark).
\State Take all training samples of the two tasks in $\mathcal C$
(32,000 in total, covering the full SNR range); peak-normalize,
compute DSP4 features (\cref{eq:dsp4}), and train the
gradient-boosted-tree specialist.
\State Calibrate the specialist threshold $\tau$ by grouped $5$-fold
out-of-fold prediction on the specialist training samples.
\Statex \textbf{\# Inference (per sample $\mathbf x$)}
\State Apply center cropping or zero-padding to $\mathbf x$ to obtain
$1{,}024$ points; peak-normalize.
\State $(\hat y_1,\hat y_2)\gets$ the top-2 candidates of
$R(\mathbf x)$.
\If{$\{\hat y_1,\hat y_2\}=\mathcal C$}
\State Compute DSP4 features; $\hat y\gets$ specialist decision
against $\tau$.
\Else
\State $\hat y\gets\hat y_1$.
\EndIf
\State \Return $(\hat y,e_{\hat y})$.
\end{algorithmic}
\end{algorithm}

\subsection{Global Decision via the Primary Classifier}
\label{sec:method:base}

The primary classifier makes a fast initial prediction over seven routing classes while
encouraging separation between task clusters in feature space. To standardize native
sequence lengths and amplitude scales across tasks, each sample is center-cropped or
zero-padded to $1{,}024$ points and then peak-normalized as $\mathbf x \leftarrow
\mathbf x/(\max\lvert\mathbf x
\rvert+10^{-6})$. This operation rescales the amplitude without subtracting the mean, preserving the second-order noncircularity information used by the subsequent physical features (see \cref{eq:dsp4}).

The normalized I/Q sequence is fed into a four-layer one-dimensional CNN to extract temporal features.
The convolutional layers have $64$, $128$, $256$, and $256$
channels, with kernel sizes $7$, $5$, $3$, and $3$, respectively.
Each block consists of a one-dimensional convolution followed by batch normalization and
ReLU activation. The first three blocks are followed by max pooling with stride $4$ to
progressively reduce temporal resolution. The fourth block has no max pooling. Its
output undergoes adaptive average pooling and linear projection to obtain a
$256$-dimensional embedding
$\mathbf h$:
\begin{equation}
\mathbf h=f_{\boldsymbol\theta}(\mathbf x)\in\mathbb R^{256},
\label{eq:cnn_embedding}
\end{equation}
where $f_{\boldsymbol\theta}$ is the convolutional feature extractor, including pooling
and linear projection, $\boldsymbol\theta$ denotes its trainable parameters, and
$\mathbf x$ is the normalized fixed-length I/Q input.

To better separate similar tasks in feature space, we apply a normalized additive-margin
classification head to the embedding $\mathbf h$~\cite{wang2018cosface}. Its trainable
weight matrix is
$\mathbf W=[\mathbf w_1,\ldots,\mathbf w_7]
\in\mathbb R^{256\times7}$, where $\mathbf w_j\in\mathbb R^{256}$
is the weight vector for routing class $j$.
For an embedding $\mathbf h$, the cosine similarity to each class weight and the
corresponding training logit are defined by
\begin{align}
\cos\theta_j
&=\frac{\mathbf h^\top\mathbf w_j}
{\lVert\mathbf h\rVert_2\lVert\mathbf w_j\rVert_2},
\notag\\
\ell_j^{\mathrm{train}}
&=s\bigl(\cos\theta_j-m\delta_{jy}\bigr),
\label{eq:margin_head}
\end{align}
where $\delta_{jy}=1$ if $j$ is the true routing class $y$, and $0$ otherwise. The scale
factor and cosine margin are set to $s=30$ and $m=0.3$, respectively.
The model is trained using standard cross-entropy loss on the logits in
\cref{eq:margin_head}, encouraging greater separation between task embeddings under the
cosine metric. At inference time, the margin term is removed, and the cosine
similarities $\{\cos\theta_j\}_{j=1}^7$ serve as the initial routing scores. If the
sample does not meet the confusion-pair trigger condition described below, its final
class is $\hat y=\arg\max_j\cos\theta_j$. Otherwise, the local re-evaluator makes a
second decision using the same sample.

\subsection{Local Re-evaluation via the Confusion-Pair Specialist}
\label{sec:method:gate}

\begin{figure}[!t]
\centering
\includegraphics[width=0.49\textwidth]{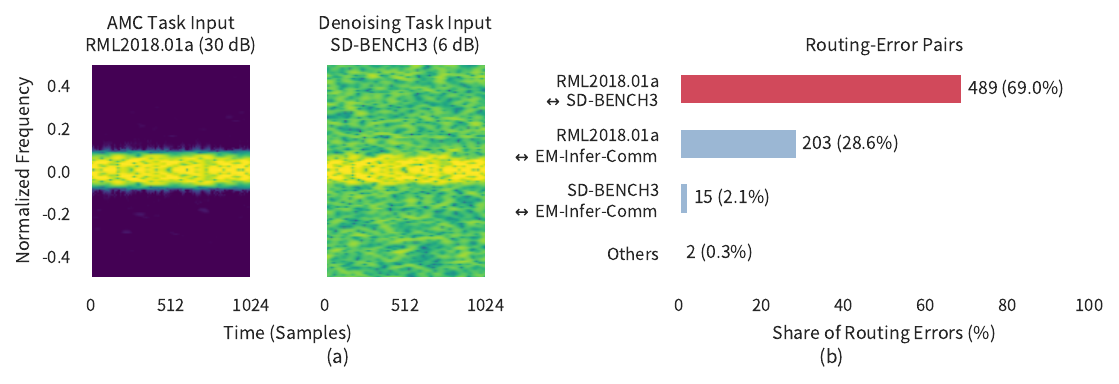}
\caption{Motivation for local re-evaluation. (a) Spectrograms of an AMC
input from RML2018.01a and a denoising input from SD-BENCH3 exhibit similar
time-frequency structures. (b) Distribution of validation routing errors over
unordered task pairs.}
\label{fig:confusion_motivation}
\end{figure}

The first-stage classifier routes most samples directly, but shared modulation patterns
or channel-induced degradation can produce similar waveforms and concentrate errors on
particular task pairs (\cref{fig:confusion_motivation}). After training, we group its
validation errors by unordered task pair and identify any pair accounting for more than
$\rho=0.5$ of all errors as a confusion pair. Pairs are therefore identified from
validation predictions rather than manually by dataset name. In our benchmark, only
$\mathcal{C}=\{\mathrm{RML2018.01a},
\mathrm{SD\mbox{-}BENCH3}\}$ meets this condition,
accounting for $69.0\%$ ($489$) of all validation routing errors.
The errors are also strongly asymmetric: among $372$
RML2018.01a samples in the $0$--$10$~dB interval of the
validation set, $371$ are misclassified as SD-BENCH3, while
there are no errors in the opposite direction.
At inference time, the local re-evaluator is invoked only when the primary classifier's
top two candidates match $\mathcal{C}$; otherwise, its top-1 prediction is used directly.

The physical differences between triggered samples may weaken under noise without
disappearing entirely. Consider second-order noncircularity.
Under zero-mean, circularly symmetric additive white Gaussian noise that is independent
of the signal, as used in both datasets, the definition of noncircularity yields
$\eta_x=\eta_s\,\gamma/(1+\gamma)$ at the population level,
where $\gamma$ is the linear signal-to-noise power ratio~\cite{schreier2010statistical}.
This relation shows that $\eta_x$ approaches the signal's noncircularity at high SNR but
decreases as SNR falls. We evaluate the discriminative ability of finite-length
estimates experimentally.
To complement noncircularity with spectral and amplitude information, we use four
physical statistics (DSP4) as input to the local re-evaluator:
\begin{align}
\eta &= \bigl\lvert \mathbb E[\mathbf z^2\bigr]\bigr\rvert\,
\big/\,\mathbb E[\lvert\mathbf z\rvert^2],
\notag\\
c &= 10^3\cdot\frac{\sum_k f_k Z_k}{\sum_k Z_k},
\notag\\
\mathrm{SFM} &= \exp\bigl(\mathbb E[\ln(Z+\epsilon)]\bigr)
\big/\,\overline Z,
\notag\\
\kappa &= \mathbb E\bigl[(a-\bar a)^4\bigr]\big/
\bigl(\mathrm{Var}(a)^2+\epsilon\bigr),\qquad a=\lvert\mathbf z\rvert.
\label{eq:dsp4}
\end{align}
The four statistics are noncircularity magnitude, spectral centroid, spectral flatness,
and amplitude kurtosis. Here, $\mathbf z$ is the peak-normalized complex I/Q signal, and
$a=\lvert\mathbf z\rvert$ denotes its amplitude. $Z_k$ is the unshifted two-sided power
spectrum formed
from the squared magnitude of the $1{,}024$-point discrete Fourier
transform, and $\overline Z$ is its mean over the full spectrum.
$f_k$ is the corresponding normalized frequency in cycles per sample.
The factor $10^3$ in the centroid provides numerical scaling only.
The numerical floor $\epsilon=10^{-12}$ prevents taking the logarithm of zero or
dividing by zero variance. The other denominators
remain nonzero because each peak-normalized sample has positive
energy. All four statistics are analytically invariant to a
global phase rotation $\mathbf z\leftarrow\mathbf z e^{j\phi}$.

\begin{figure}[!t]
\centering
\resizebox{0.49\textwidth}{!}{\includegraphics{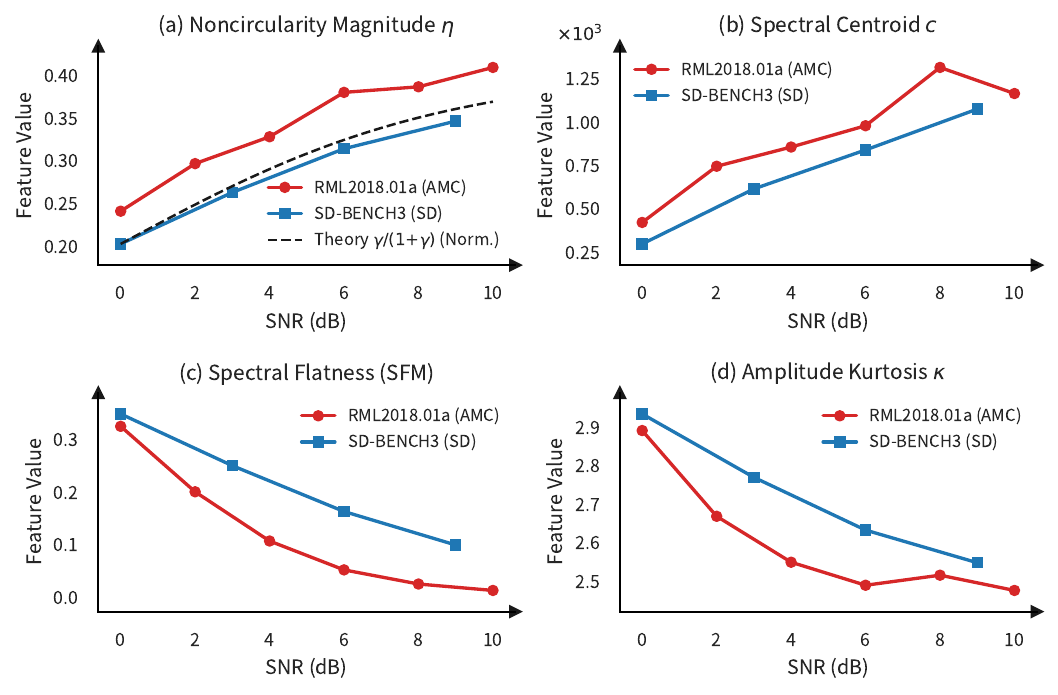}}
\caption{Means of the four DSP4 statistics on the training splits,
stratified by class and SNR tier over the 0--10~dB confusion region.
The tiers follow the native SNR grids of the two datasets
(RML2018.01a: even-numbered tiers; SD-BENCH3: 0/3/6/9~dB). The class
means differ at every shown tier; the dashed curve in (a) is the
theoretical attenuation law $\gamma/(1{+}\gamma)$ normalized to the
SD mean at 0~dB.}
\label{fig:dsp4phys}
\end{figure}

Three sets of experiments support the discriminative value of these statistics. First,
classifiers trained separately at each SNR level show that the physical statistics can
distinguish the two signal classes under the same noise conditions. At SNR levels shared
by both datasets, we train separate binary classifiers on DSP4 features using the same
gradient-boosting model type as the local re-evaluator. Balanced
accuracy reaches $74.2\%$ at $0$~dB and $100\%$ at $6$~dB.
Second, the class means of the four statistics differ at every examined SNR level in the confusion interval
(\cref{fig:dsp4phys}), supporting the physical basis for local
re-evaluation. Third, counterfactual probes show that the local re-evaluator responds to
the tested impairment processes, even on previously unseen test samples
(Supplementary Material, Section~D).

We use histogram-based gradient boosting (HGBT) to classify the four standardized physical
statistics, allowing nonlinear relations and interactions among features. HGBT can
fit complex decision boundaries in a low-dimensional space
using an ensemble of trees, without assuming a particular
distribution for the statistics. Its low training and inference costs also suit a
lightweight local re-evaluator. The training set contains all samples
from the training splits of the two datasets and covers the
full SNR range. We calibrate the decision threshold on the training set using grouped
five-fold out-of-fold predictions and the Youden index~\cite{youden1950index}.
The folds preserve class proportions, and samples sharing
the same source are assigned to the same fold. The grouping
rule is detailed in Section~C of the Supplementary Material.
Each fold and the final model follow the same training rules.
The validation and test splits are not used to train the local re-evaluator or calibrate its threshold.
At inference time, the local re-evaluator uses only physical statistics of the current
sample, without requiring SNR labels or a noise estimation module.

\section{Experiments}
\label{sec:Experiments}

\subsection{Implementation Details}
\label{sec:impl}

The primary classifier and end-to-end baselines are trained for $20$ epochs using Adam
with an initial learning rate of $10^{-3}$, weight decay of $10^{-4}$, a batch size of
$512$, and cosine learning-rate annealing. Checkpoints are selected by the highest
accuracy on the validation set.
Each expert is trained separately and selected using its
task-specific metric: validation overall accuracy for
classification experts, validation SNR gain for the SD
expert, and validation SI-SNR loss for the BSS expert.
We train the primary classifier with three random seeds, varying both weight
initialization and data order, to assess performance variation.
Each baseline is trained once.

The $19$ baselines comprise three end-to-end deep models, two frozen-feature
classifiers, three routing methods, and $11$ classical machine learning classifiers.
An Oracle with perfect routing provides an upper bound and is not counted among the
baselines. The training settings
for the four groups are described below.

All non-end-to-end baselines use a shared front end with frozen AWN features. We select
the feature layer using a two-step linear-probing procedure. The candidate backbones are
three modulation recognition networks: AWN, IQFormer, and AMC-Net. Each candidate
backbone is trained on its corresponding modulation dataset, and the best checkpoint is
selected (AWN-2018a denotes AWN trained on RML2018.01a).
First, we train identical logistic regression probes on the penultimate-layer features
of each backbone to predict the task label. AWN-2018a achieves $89.18\%$ accuracy,
compared with $72.18\%$ for IQFormer and $70.68\%$ for AMC-Net.
Second, we compare the layers of AWN-2018a. Under the final
split protocol, the \texttt{se\_out} layer has the highest
single-layer probe accuracy ($92.40\%$), and is therefore used as the routing feature.
Both the feature-based and classical
machine learning baselines use this front end.
In contrast, AMC-Net~\cite{zhang2023amc}, ResNet~\cite{o2018over}, and
PET-CGDNN~\cite{zhang2021efficient} take unified-length I/Q samples as input and are
retrained using the same optimizer and number of epochs.
The three routing baselines retain the main mechanisms of their official
implementations, are adapted to seven classes, and use the same frozen AWN front end.
CSCR~\cite{shirkavand2026cost} trains a contrastive encoder
to align sample embeddings with frozen expert behavior
fingerprints, and selects an expert by nearest-neighbor retrieval at inference time.
EmbedLLM~\cite{zhuang2025embedllm} trains a
correctness prediction head for each expert and uses the resulting scores for routing.
GraphRouter~\cite{feng2410graphrouter} uses a
graph neural network to learn selection distributions on a
sample--expert graph. All three baselines and GLCR use the same training and validation splits.

Training is conducted on an NVIDIA V100 GPU using Python 3.10, PyTorch 2.5.1, and CUDA 12.1.
Routing latency is measured in FP32 on a local machine's CPU (AMD Ryzen 9 9955HX, $32$ threads).
The measured latency covers the complete routing pipeline, from a $2\times1{,}024$ I/Q
sample to the final routing decision. The cost of the physical-statistics
branch is weighted by its measured trigger rate on the test
set ($18.0\%$). After warm-up, we measure $200$ runs at batch size $1$ and report the
mean per-sample latency. The source code and lightweight reproduction assets are publicly
available at \url{https://github.com/Dawn-try/GLCR}.

\subsection{Results and Analysis}
\label{sec:results}

\cref{tab:routing_acc} reports the routing accuracy and system
score of all methods on the HeteroSignal-7 test split.
The system score measures normalized task utility: a misrouted sample receives zero, a
correctly routed classification sample is scored by prediction correctness, and
correctly routed SD and BSS samples are scored using
$\mathrm{clip}(\Delta\mathrm{SNR}/20,0,1)$ and
$\mathrm{clip}(\Delta\mathrm{SI\text{-}SNR}/20,0,1)$,
respectively. Here, $\Delta\mathrm{SNR}$ denotes the SNR gain over the noisy input, and
$\Delta\mathrm{SI\text{-}SNR}$ denotes the SI-SNR improvement over the mixture after
optimal permutation matching. The $20$~dB normalization scale was fixed during benchmark
design, is shared by all methods, and does not affect routing decisions.

GLCR achieves $96.16\%$ routing accuracy and a $76.30\%$ system score, both the highest
among the compared methods.
Compared with the strongest signal model,
AMC-Net~\cite{zhang2023amc} ($95.01\%$/$74.89\%$),
GLCR improves routing accuracy by $+1.16\%$
and system score by $+1.41\%$.
Its primary classifier has $0.41$M parameters, approximately one-eleventh of AMC-Net's $4.60$M.
ResNet achieves the second-highest system score ($75.16\%$); GLCR exceeds this score by $1.14\%$.
To account for possible correlation among SD samples derived from the same clean
reference, we use a task-stratified paired cluster bootstrap to compare GLCR and
AMC-Net. The $977$ SD reference groups are resampled as clusters, while samples from the
other tasks are resampled individually.
We compute percentile-based $95\%$ confidence intervals from $10{,}000$ resamples with
random seed $2024$, using the same resampled units for both methods. The confidence
interval for the system-score difference is
$[+1.14, +1.69]\%$, and that for
routing accuracy is $[+0.79, +1.52]\%$.
Neither interval includes zero. As a supplementary check, an exact two-sided McNemar
test on paired sample-level predictions ($b=431$, $c=269$) gives
$p=9.8\times10^{-10}$, consistent with the bootstrap
intervals. We also compare GLCR with ResNet, the second-best baseline in system score,
using the same paired testing protocol. The confidence interval
is $[+0.90, +1.39]\%$ for system score and
$[+1.02, +1.66]\%$ for routing accuracy.
The McNemar test ($b=381$, $c=194$) gives
$p=5.0\times10^{-15}$, providing further evidence of a significant difference.

On the five-dataset subset without padding, GLCR and AMC-Net achieve routing accuracies of $94.63\%$
and $93.00\%$, respectively, showing that the advantage remains when padding-related cues are absent.
The system-score gain varies across tasks. The largest gain occurs on RML2018.01a, where
the joint routing-and-expert success rate is
$52.26\%$ versus $40.85\%$, an improvement of $+11.41\%$.
On SD and WII, GLCR shows small decreases of $0.87\%$ and $0.70\%$, respectively,
while differences on the other four tasks are all within
$0.2\%$. Among classical machine learning methods, random forest
performs best, with a routing accuracy of $93.70\%$. The three recent routing methods
(CSCR~\cite{shirkavand2026cost},
EmbedLLM~\cite{zhuang2025embedllm}, and
GraphRouter~\cite{feng2410graphrouter}) achieve
$90.24\%$--$93.83\%$ routing accuracy on our benchmark,
all below AMC-Net, the strongest signal-model baseline
($95.01\%$). The perfect-routing Oracle achieves a system-score upper bound of
$78.46\%$, leaving a gap of $2.16\%$ for GLCR.

\begin{table}[!t]
\centering
\caption{Routing Accuracy and System Score on the Cleaned HeteroSignal-7
Test Split (13{,}975 Samples; Seed 0). Oracle is the upper bound and is
excluded from comparison. Best and second-best results are in \textbf{bold}
and \underline{underlined}, respectively.}
\label{tab:routing_acc}
\footnotesize
\setlength{\tabcolsep}{2.5pt}
\renewcommand{\arraystretch}{1.15}
\resizebox{\columnwidth}{!}{%
\begin{tabular}{@{}llcc@{}}
\toprule
Category & Method & \shortstack{Routing\\Acc. (\%)} & \shortstack{System\\Score (\%)} \\
\midrule
\multirow{3}{*}{\shortstack[l]{Signal\\Models}}
 & AMC-Net & \underline{95.01} & 74.89 \\
 & ResNet & 94.83 & \underline{75.16} \\
 & PET-CGDNN & 92.14 & 73.83 \\
\midrule
\multirow{2}{*}{\shortstack[l]{Feature-based\\Classifiers}}
 & \shortstack[l]{All-layer concatenation\\+ MLP} & 94.19 & 74.86 \\
 & \texttt{se\_out} + MLP & 94.03 & 74.73 \\
\midrule
\multirow{3}{*}{\shortstack[l]{Routing\\Methods}}
 & AWN + CSCR & 93.83 & 74.67 \\
 & AWN + EmbedLLM & 90.40 & 73.68 \\
 & AWN + GraphRouter & 90.24 & 73.75 \\
\midrule
\multirow{11}{*}{Classic ML}
 & Random forest & 93.70 & 74.50 \\
 & \shortstack[l]{Histogram-based\\gradient boosting} & 93.55 & 74.44 \\
 & Logistic regression & 92.96 & 74.10 \\
 & RBF-SVM & 92.93 & 73.89 \\
 & Linear SVM & 92.72 & 73.87 \\
 & $k$NN & 91.57 & 73.33 \\
 & LDA & 91.03 & 72.85 \\
 & QDA & 88.59 & 72.93 \\
 & Nearest class mean & 79.22 & 62.35 \\
 & Mahalanobis & 76.92 & 63.83 \\
 & PCA subspace & 74.20 & 61.32 \\
\midrule
\textbf{Proposed} & \textbf{GLCR}
 & \textbf{96.16} & \textbf{76.30} \\
\midrule
Upper Bound & Oracle & 100.00 & 78.46 \\
\bottomrule
\end{tabular}%
}
\end{table}

To assess training variability, we retrain GLCR with three random seeds that change both
weight initialization and data order.
The routing accuracies on the test set are $96.16\%$,
$96.09\%$, and $95.91\%$, with a mean of $96.05\%$
and a standard deviation of $0.13\%$.
The corresponding system scores are $76.30\%$,
$76.27\%$, and $76.00\%$, indicating limited variation across the three seeds.

For efficiency analysis, we model the average latency of GLCR's two-stage routing
pipeline. Let $p_{\mathrm{pair}}$ denote the probability that the top two candidates
match the confusion pair. The average
per-sample latency is
\begin{equation}
\bar T = T_{\mathrm{CNN}}
+ p_{\mathrm{pair}}\,\bigl(T_{\mathrm{feat}}+T_{\mathrm{spec}}\bigr).
\label{eq:latency}
\end{equation}
Here, $T_{\mathrm{CNN}}$ is the primary classifier's latency, $T_{\mathrm{feat}}$ is the
time needed to compute the four physical statistics (including the FFT and spectral
statistics), and $T_{\mathrm{spec}}$ is the inference cost
of the tree-based local re-evaluator (see \cref{sec:impl}
for the latency measurement protocol).
On the HeteroSignal-7 test split,
$p_{\mathrm{pair}}=0.180$: $82.0\%$ of samples are routed
with only one global forward pass, and the remaining $18.0\%$ invoke the second stage.

The primary CNN requires $42.0$M FLOPs per sample and has $0.41$M parameters; the
$300$-tree specialist occupies approximately $1.08$~MB. The mean routing-pipeline CPU
latency is $1.43$~ms per I/Q sample, including DSP4 and tree computation weighted by the
measured trigger rate but excluding expert execution. Both decision-stage FLOPs and
routing latency are the lowest among the compared methods. AMC-Net, the most accurate
signal model, requires $276.6$M FLOPs per sample; GLCR's primary classifier uses $15.2\%$
of this amount and about one quarter of its latency.

The expected computation of the selected expert depends on each method's routing distribution:
$\mathbb E[C_{\mathrm{expert}}]
=\sum_j P(\hat y=j)\,C_{e_j}$. Under the actual GLCR
routing distribution, this is $235.6$M FLOPs per sample
($241.4$M for AMC-Net), close to the $233.0$M obtained
under a uniform distribution across seven tasks.
Executing all seven experts for every sample requires at least $1{,}631$M FLOPs, whereas
the approximate expected cost of GLCR and its selected expert is
$42.0+235.6=277.6$M FLOPs per sample, a $5.9$-fold reduction.

\begin{figure}[!t]
\centering
\includegraphics[width=0.49\textwidth]{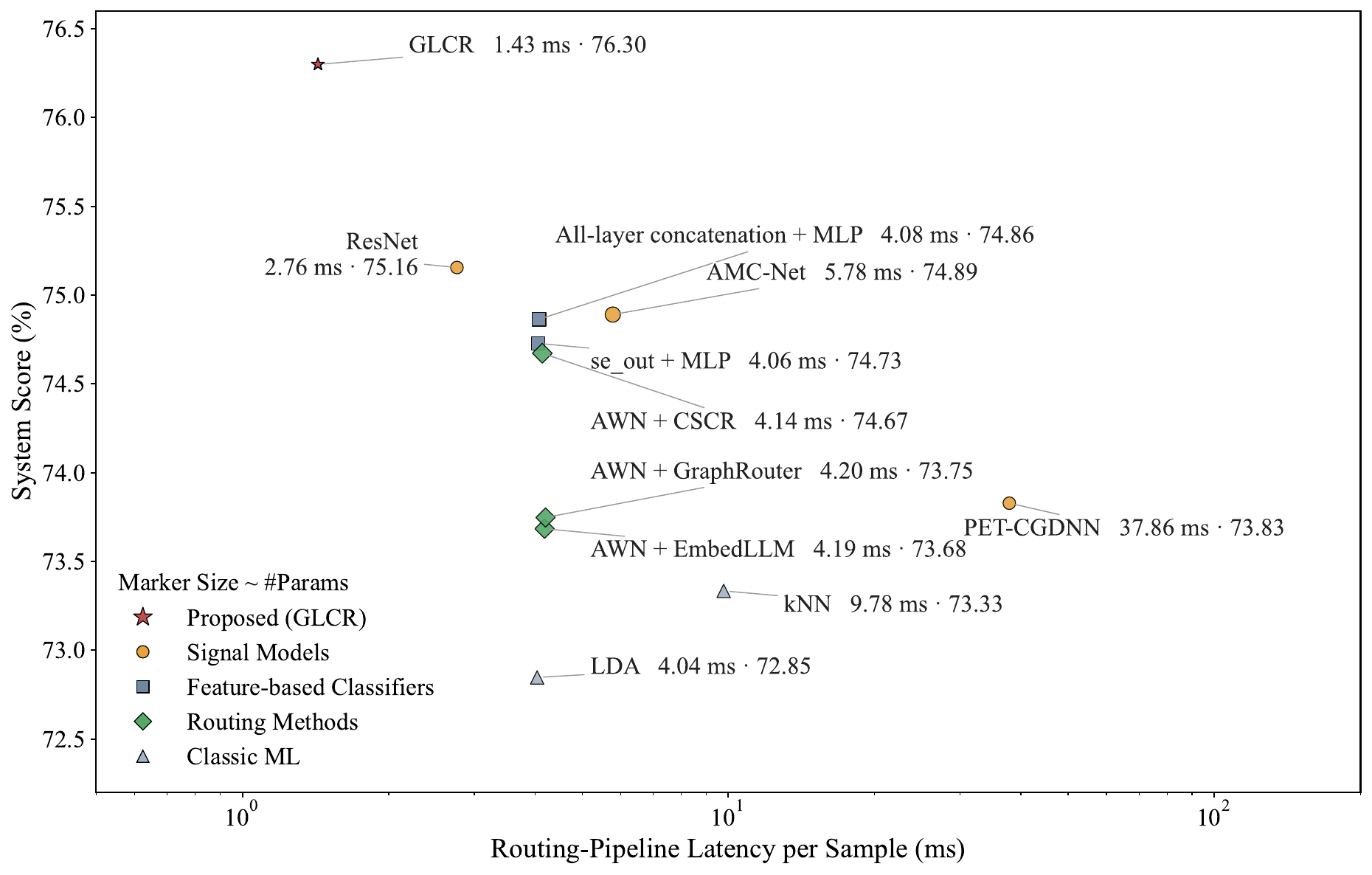}
\caption{System score on the cleaned test split versus
routing-pipeline CPU latency per sample for the compared routers;
a representative subset of the nineteen
baselines is shown for clarity.}
\label{fig:pareto}
\end{figure}

\cref{fig:pareto} shows that GLCR achieves the highest system score and lowest measured
latency, forming the only Pareto-optimal point. This conclusion covers all baselines in
\cref{tab:routing_acc}: the shared AWN prefix imposes a $4.04$~ms lower bound on
frozen-feature latency, while measured signal-model latency is at least $2.76$~ms.

\begin{figure*}[!t]
\centering
\resizebox{0.90\textwidth}{!}{%
\begin{minipage}[c]{0.42\textwidth}
\centering
\includegraphics[width=\textwidth]{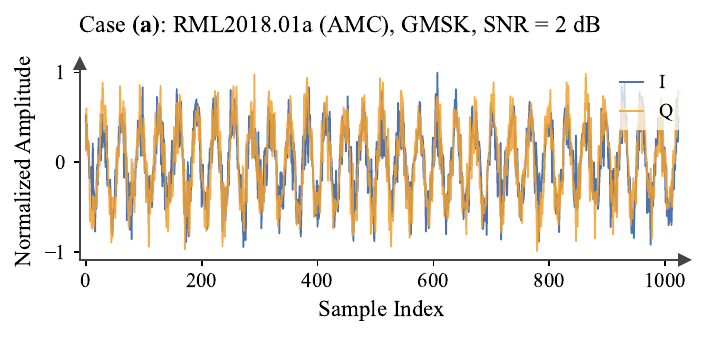}\\[2mm]
\includegraphics[width=\textwidth]{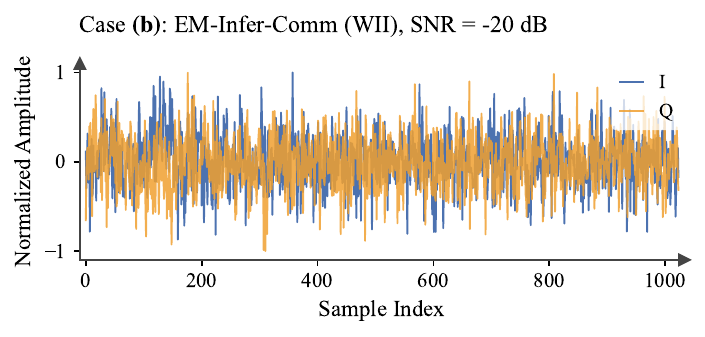}
\end{minipage}\hfill
\begin{minipage}[c]{0.55\textwidth}
\centering
\footnotesize
\renewcommand{\arraystretch}{1.35}
\begin{tabular}{@{}cllc@{}}
\toprule
Case & Model & Routed To & Correct \\
\midrule
\multirow{6}{*}{(a)}
 & kNN & SD-BENCH3 & $\times$ \\
 & All-layer concatenation + MLP & SD-BENCH3 & $\times$ \\
 & AWN + CSCR & SD-BENCH3 & $\times$ \\
 & AWN + GraphRouter & SD-BENCH3 & $\times$ \\
 & AMC-Net & SD-BENCH3 & $\times$ \\
 & \textbf{GLCR} & \textbf{RML2018.01a} & \textbf{$\checkmark$} \\
\midrule
\multirow{6}{*}{(b)}
 & kNN & RML2018.01a & $\times$ \\
 & All-layer concatenation + MLP & RML2018.01a & $\times$ \\
 & AWN + CSCR & RML2018.01a & $\times$ \\
 & AWN + GraphRouter & RML2018.01a & $\times$ \\
 & AMC-Net & RML2018.01a & $\times$ \\
 & \textbf{GLCR} & \textbf{EM-Infer-Comm} & \textbf{$\checkmark$} \\
\bottomrule
\end{tabular}
\end{minipage}%
}
\caption{I/Q waveforms and routing decisions of the compared
methods for two case-study samples.}
\label{fig:case_study}
\end{figure*}

\cref{fig:case_study} illustrates two difficult routing decisions. Case (a) is a GMSK signal from
RML2018.01a (SNR $=2$~dB), and case (b) is an interference
signal from EM-Infer-Comm (SNR $=-20$~dB).
In case (a), all five representative baselines in
\cref{fig:case_study} misroute the sample to the denoising
expert (SD-BENCH3). In case (b), all five baselines
incorrectly predict RML2018.01a at $-20$~dB.
GLCR is the only method shown that routes both samples correctly.
Case (a) illustrates the intended role of the local re-evaluator: the CNN backbone's top-2 candidates
are RML2018.01a and SD-BENCH3, triggering a second
decision based on physical statistics
($p=0.82\geq0.535$), which corrects the route.
In case (b), the primary CNN predicts the correct class directly, without invoking the specialist.

\subsection{Ablation Study}
\label{sec:ablation}

\begin{table*}[!t]
\centering
\caption{Ablation study of GLCR on HeteroSignal-7 (\checkmark: enabled). $\Delta$ denotes routing accuracy difference relative to the preceding row in Panel~A, the full configuration in Panel~B, and HGBT in Panel~C.}
\label{tab:ablation}
\footnotesize
\renewcommand{\arraystretch}{1.2}
\begin{tabular}{@{}lccccccc@{}}
\toprule
 & \multicolumn{2}{c}{Stage-1 classifier}
 & \multicolumn{2}{c}{Stage-2 specialist} & \multicolumn{3}{c}{Results} \\
\cmidrule(lr){2-3} \cmidrule(lr){4-5} \cmidrule(l){6-8}
Configuration & CNN & Margin & Input & Classifier
 & Acc. (\%) & Sys. (\%) & $\Delta$ (\%) \\
\midrule
\multicolumn{8}{@{}l}{\emph{Panel A: construction path}}\\
\quad MLP on frozen AWN features (\texttt{se\_out})
 &  &  &  &
 & 94.03 & 74.73 & -- \\
\quad End-to-end CNN with a plain linear head
 & $\checkmark$ &  &  &
 & 94.74 & 75.22 & $+0.71$ \\
\quad \enspace + feature \& weight normalization ($m{=}0$)
 & $\checkmark$ &  &  &
 & 94.54 & 75.17 & $-0.20$ \\
\quad \enspace + additive margin $m{=}0.3$ (stage-1 classifier)
 & $\checkmark$ & $\checkmark$ &  &
 & 94.91 & 75.34 & $+0.37$ \\
\quad \enspace + specialist on the confusion pair (\textbf{GLCR})
 & $\checkmark$ & $\checkmark$ & DSP4 & HGBT
 & \textbf{96.16} & \textbf{76.30} & $+1.25$ \\
\midrule
\multicolumn{8}{@{}l}{\emph{Panel B: specialist design}}\\
\quad GLCR (full configuration)
 & $\checkmark$ & $\checkmark$ & DSP4 & HGBT
 & 96.16 & 76.30 & -- \\
\quad Stage-1 classifier only (no specialist)
 & $\checkmark$ & $\checkmark$ & -- & --
 & 94.91 & 75.34 & $-1.25$ \\
\quad Specialist input: raw I/Q (2{,}048-D)
 & $\checkmark$ & $\checkmark$ & \emph{raw I/Q} & HGBT
 & 93.46 & 74.28 & $-2.70$ \\
\quad Specialist input: CNN embedding (256-D)
 & $\checkmark$ & $\checkmark$ & \emph{CNN emb.} & HGBT
 & 94.91 & 75.37 & $-1.25$ \\
\quad Specialist input: generic statistics (RAW8)
 & $\checkmark$ & $\checkmark$ & \emph{RAW8} & HGBT
 & 94.53 & 75.12 & $-1.63$ \\
\quad Plain CE head with the same specialist
 & $\checkmark$ &  & DSP4 & HGBT
 & 96.06 & 76.29 & $-0.10$ \\
\quad End-to-end fusion (single CNN, DSP4 logit offset)
 & $\checkmark$ & $\checkmark$ & \emph{fused} & --
 & 94.83 & 75.35 & $-1.33$ \\
\midrule
\multicolumn{8}{@{}l}{\emph{Panel C: specialist classifier on DSP4
(mean of three seeds)}}\\
\quad HGBT (1.05~MB, 14~$\mu$s)
 & $\checkmark$ & $\checkmark$ & DSP4 & HGBT
 & 96.05 & 76.19 & -- \\
\quad Random forest (100.7~MB, 125~$\mu$s)
 & $\checkmark$ & $\checkmark$ & DSP4 & \emph{RF}
 & 96.06 & 76.19 & $+0.01$ \\
\quad kNN (2.2~MB, 51~$\mu$s)
 & $\checkmark$ & $\checkmark$ & DSP4 & \emph{kNN}
 & 95.14 & 75.44 & $-0.91$ \\
\quad RBF-SVM (0.3~MB, 195~$\mu$s)
 & $\checkmark$ & $\checkmark$ & DSP4 & \emph{SVM}
 & 94.41 & 74.85 & $-1.64$ \\
\quad MLP (0.04~MB, 0.4~$\mu$s)
 & $\checkmark$ & $\checkmark$ & DSP4 & \emph{MLP}
 & 94.58 & 75.00 & $-1.47$ \\
\quad Logistic regression ($<$0.01~MB, 0.1~$\mu$s)
 & $\checkmark$ & $\checkmark$ & DSP4 & \emph{LR}
 & 94.15 & 74.71 & $-1.90$ \\
\bottomrule
\end{tabular}
\end{table*}

Panel A of \cref{tab:ablation} examines the contributions of successive components of GLCR.
Replacing the frozen-feature MLP with an end-to-end CNN
increases routing accuracy from $94.03\%$ to $94.74\%$
($+0.71\%$). The effect of the classification head can be examined in
two steps. Applying feature and weight normalization without a margin ($m{=}0$) lowers
accuracy to $94.54\%$
($-0.20\%$). Normalization favors RML2018.01a, increasing its recall by $3.7\%$
while reducing SD-BENCH3 recall by $4.4\%$. Adding the margin
($m{=}0.3$) raises accuracy to $94.91\%$ ($+0.37\%$), with smaller changes in both
recalls. Relative to the plain linear head, RML2018.01a recall increases from $0.703$ to
$0.721$, while SD-BENCH3 recall remains nearly unchanged
($0.993\rightarrow0.990$).

Normalization and the additive margin have complementary effects. Without normalization,
the same absolute logit margin ($m \cdot s = 9$) yields only a $+0.05\%$ gain
because the unnormalized logits can grow in magnitude (the mean target-class logit
is about $56$), reducing the effect of a fixed margin.
Normalization maps embeddings and class weights onto the unit sphere, making the logits
depend on cosine similarity rather than feature magnitude. The margin thus corresponds
to a fixed offset in cosine space.
Adding the second stage, a local re-evaluator trained over the full SNR range and
triggered by the top-two confusion pair, completes GLCR. Routing
accuracy increases to $96.16\%$ ($+1.25\%$), and
system score rises from $75.34\%$ to $76.30\%$.

Balanced sample weighting in the $0$--$10$~dB confusion interval instead reduces
accuracy: the cross-entropy head drops from $94.74\%$ to $94.58\%$, and the margin head
from $94.91\%$ to $94.53\%$. With the margin head, RML2018.01a recall falls from $0.721$
to $0.714$ and SD-BENCH3 recall from $0.990$ to $0.965$, so neither target class
improves. Shared parameters can shift the decision boundary both inside and outside the
reweighted interval; thus, the net effect cannot be inferred from the class-prior
adjustment alone~\cite{menon2020long}. This negative result motivates \emph{post-hoc},
sample-triggered local re-evaluation rather than training-data reweighting.

Panel B evaluates design choices for the local re-evaluator. Each design is selected on
the validation set and evaluated, without further tuning, on the cleaned test set.
$\Delta$ denotes the difference from the full configuration.
For a fair comparison, each input and classifier variant has its own threshold
calibrated by grouped out-of-fold predictions and the Youden index. Feature scaling is
fitted only on the training portion of each fold.

DSP4 outperforms all three alternative inputs in the ablation study. Using raw I/Q
samples ($2{,}048$ dimensions) as the specialist input lowers routing accuracy to
$93.46\%$ ($-2.70\%$).
Using the primary CNN embedding ($256$ dimensions) yields $94.91\%$ ($-1.25\%$).
Using eight generic I/Q statistics (RAW8: mean,
standard deviation, skewness, and kurtosis of amplitude;
standard deviations of I and Q; average power; and
standard deviation of instantaneous phase differences)
gives $94.53\%$ ($-1.63\%$). DSP4 provides global statistical features that a tree model
may struggle to learn directly from raw I/Q samples under the tested capacity and
training budget. These features complement the CNN embedding.

Replacing the margin head with a standard cross-entropy head, while retaining the specialist, yields
$96.06\%$ ($-0.10\%$), indicating that the margin head remains useful when local re-evaluation is enabled.
Finally, an end-to-end fusion variant, which adds
DSP4-based offsets to the two confusion-pair logits
within the same network, achieves only $94.83\%$
($-1.33\%$). Its RML2018.01a recall is near zero in the $0$--$10$~dB interval, indicating
that this tested fusion design fails to correct the confusion-pair errors using physical
statistics. The two-stage configuration
outperforms the fusion variant tested here.
Routing across all tasks using DSP4 alone, without a CNN, achieves only $85.65\%$. Thus,
in this setting, physical statistics do not replace features learned from signal
content.

Panel C compares alternative classifiers for the specialist. Results are averaged over
three seeds; parentheses report serialized model size and per-sample CPU latency. We use
the same CPU as in the main latency measurements: after warm-up on $512$ samples, we run
$30$ batch inferences and divide the elapsed time by the number of samples to obtain
per-sample latency in microseconds. Tree ensembles achieve higher accuracy than the
other tested classifiers. HGBT and random forest achieve nearly identical accuracy
($96.05\%$ and $96.06\%$), differing by much less than the across-seed standard
deviation of $0.13\%$. However, random forest is about two orders of
magnitude larger and nearly one order of magnitude slower. kNN, SVM, MLP, and logistic
regression are all $0.9\%$--$1.9\%$ lower. HGBT therefore provides the best
accuracy--cost trade-off and is selected.

\FloatBarrier
\section{Conclusion}
\label{sec:Conclusion}

This paper investigates routing to a heterogeneous pool of electromagnetic signal
experts. We construct HeteroSignal-7, comprising six task types, seven datasets, and
$140{,}000$~samples, paired with a heterogeneous pool of seven classification and
waveform-recovery experts under a unified evaluation protocol. We also propose GLCR, a
two-stage routing method. A lightweight CNN with an additive-margin head predicts one of
seven routing classes. When the top two candidates are RML2018.01a and SD-BENCH3, a local
re-evaluator uses four physical statistics to make a second decision. GLCR requires
neither SNR labels nor a noise estimation module at inference time. With this lightweight
design, extensive experiments and ablation studies on HeteroSignal-7 demonstrate that
GLCR consistently outperforms all $19$ baselines in both routing accuracy and system
score while achieving the lowest routing latency and computational cost.

\textbf{Limitations.} (i) GLCR remains $2.27\%$
below the perfect-routing upper bound in
system score, based on the three-seed mean and
Oracle upper bound. Many remaining errors are concentrated on the RML2018.01a--SD-BENCH3
confusion pair, suggesting that the four statistics do not
capture all physical differences between the two
signal classes. Richer physical statistics or a
lightweight time-frequency branch may further
reduce this gap. (ii) The current local re-evaluator
handles only one confusion pair. The trigger-and-specialist design could be extended to
handle multiple confusion pairs. (iii) Our protocol assumes
closed-set routing; handling unknown tasks through open-set routing remains future work.
(iv) Only one dataset in the benchmark comes from
real-world measurements. We plan to include more measured signal sources to evaluate the
protocol in broader settings.

\bibliographystyle{IEEEtran}
\bibliography{reference}

\raggedbottom
\makeatletter
\def\@IEEEBIOskipN{0pt}
\makeatother

\begin{IEEEbiography}[{\includegraphics[width=1in,height=1.25in,clip,keepaspectratio]{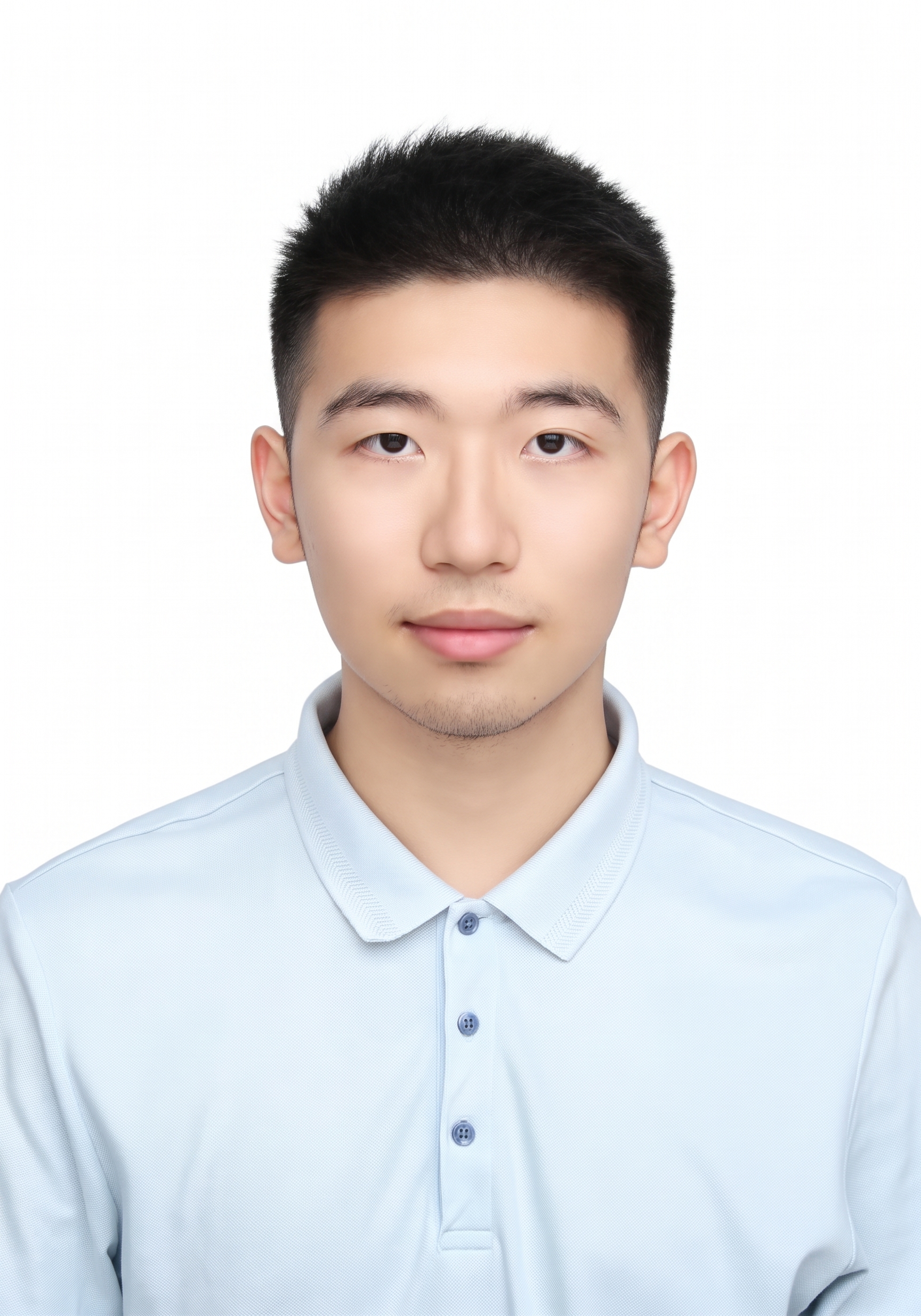}}]{Fang Lu}
is currently pursuing the B.S. degree in intelligent science and technology with the Institute of Cyberspace Security, College of Information Engineering, and Jianxing Honors College, Zhejiang University of Technology, Hangzhou, China, where he is also with the Binjiang Institute of Artificial Intelligence.

His current research interests include electromagnetic signal processing, multi-task signal routing, signal denoising, and physics-aware deep learning.
\end{IEEEbiography}
\begin{IEEEbiography}[{\includegraphics[width=1in,height=1.25in,clip,keepaspectratio]{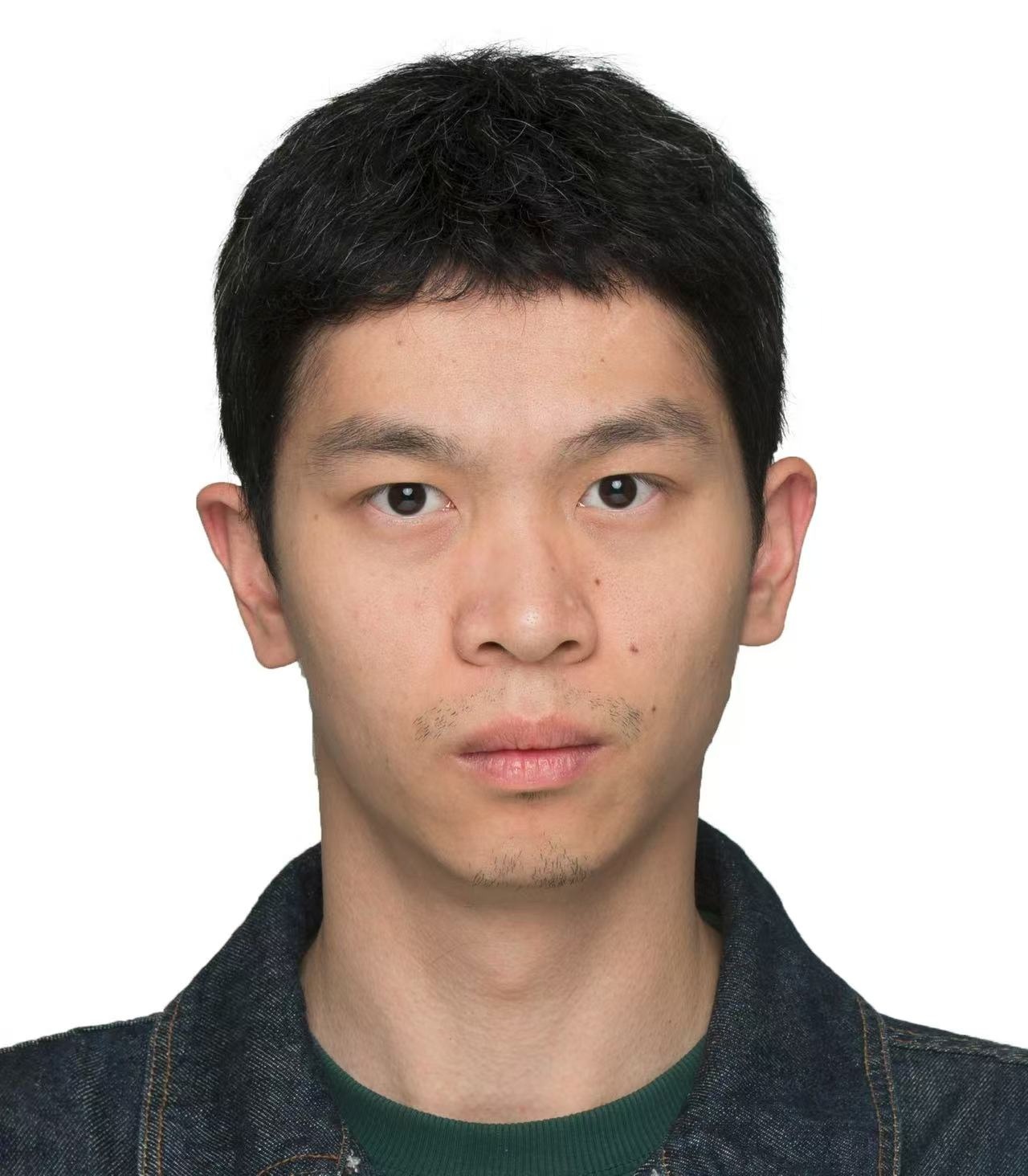}}]{Yao Lu}
received his B.S. degree and Ph.D. degree in control science and engineering from
Zhejiang University of Technology. He is currently a postdoctoral researcher at Zhejiang
University of Technology, Hangzhou, China. He is a visiting scholar with the Centre for
Frontier AI Research, Agency for Science, Technology and Research, Singapore. He has
published papers in international conferences and journals, including ECCV, ICLR, AAAI,
TNNLS, Neurocomputing, and TCCN. He serves as a reviewer for ICLR, CVPR, ICCV, ECCV,
ICML, AAAI, NeurIPS, IoT-J, and TNNLS. His research interests include deep learning and
computer vision, with a focus on artificial intelligence and model compression.
\end{IEEEbiography}
\begin{IEEEbiography}[{\includegraphics[width=1in,height=1.25in,clip,keepaspectratio]{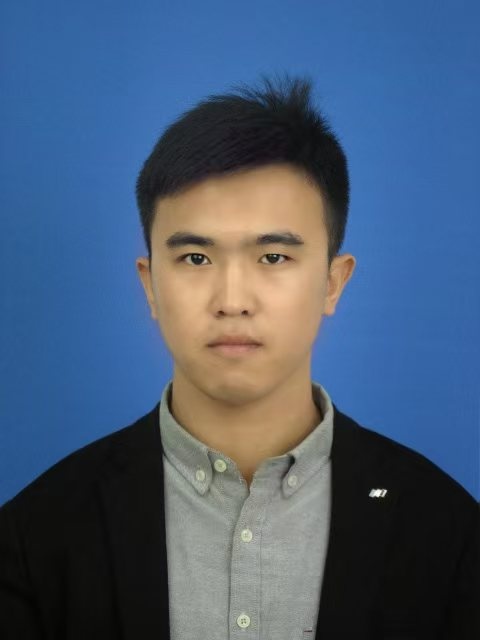}}]{Zhuangzhi Chen}
received the B.S. and Ph.D. degrees in Control Theory and Engineering from Zhejiang
University of Technology, Hangzhou, China, in 2017 and 2022, respectively. He was a
visiting scholar with the Department of Computer Science, University of California at
Davis, CA, in 2019. He completed his postdoctoral research in Computer Science and
Technology at Zhejiang University of Technology, in 2024. He is currently an associate
researcher at the Cyberspace Security Research Institute of Zhejiang University of
Technology and deputy director and associate researcher at the Research Center of
Electromagnetic Space Security, Binjiang Institute of Artificial Intelligence, ZJUT,
Hangzhou 310056, China.

His research interests include deep learning algorithms, deep learning-based radio
signal recognition, and AI for wireless communication security.
\end{IEEEbiography}
\begin{IEEEbiography}[{\includegraphics[width=1in,height=1.25in,clip,keepaspectratio]{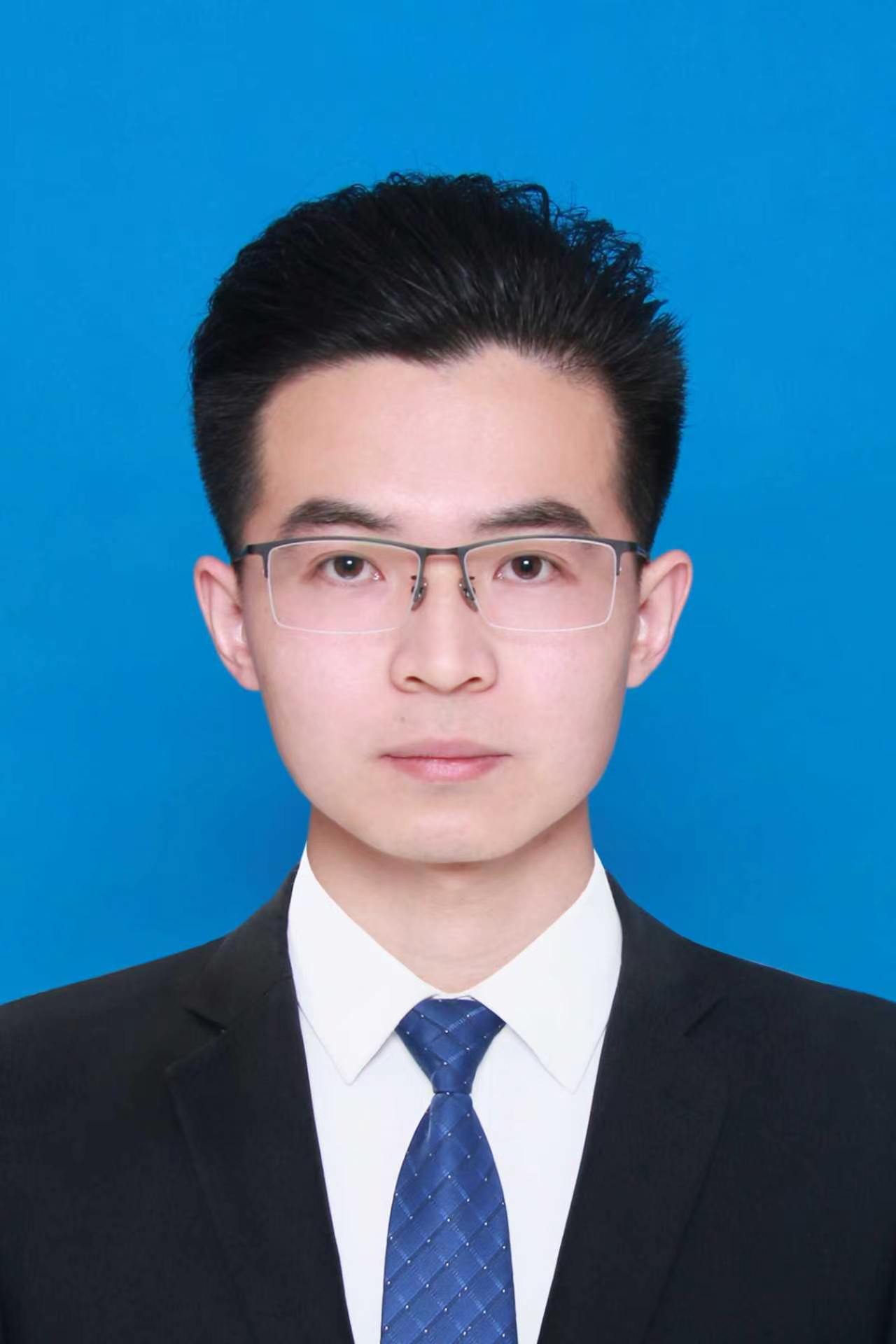}}]{Sicheng Zhang}
(Member, IEEE) received the BS and PhD degrees from Harbin Engineering University, Harbin, China, in 2019 and 2025, respectively. He is currently a faculty-track postdoc with the College of Information and Communication Engineering, Harbin Engineering University. His research interests include communication technology, signal processing, artificial intelligence and adversarial threat in electromagnetic space and lightweight deep learning model. He served as publication chair of ADHIP 2023. He serves as an editorial board member of the 2026 Academic Conference on Electromagnetic Spectrum. He has gotten the best paper awards of ICCC 2023, WiSATS 2025, and MOBIMEDIA 2025.
\end{IEEEbiography}
\begin{IEEEbiography}[{\includegraphics[width=1in,height=1.25in,clip,keepaspectratio]{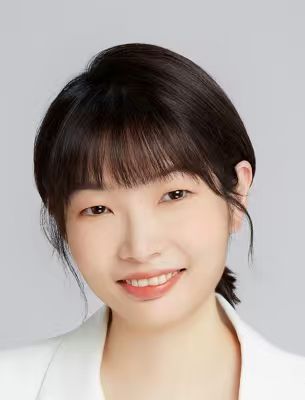}}]{Shanqing Yu}
received an M.S. degree from the School of Computer Engineering and Science, Shanghai
University, China, in 2008, an M.S. degree from the Graduate School of Information,
Production, and Systems, Waseda University, Japan, in 2008, and a Ph.D. degree in 2011.
She is currently an Associate Professor at the Institute of Cyberspace Security and the
College of Information Engineering, Zhejiang University of Technology, Hangzhou, China.
Her research interests cover intelligent computation and data mining.
\end{IEEEbiography}
\begin{IEEEbiography}[{\includegraphics[width=1in,height=1.25in,clip,keepaspectratio]{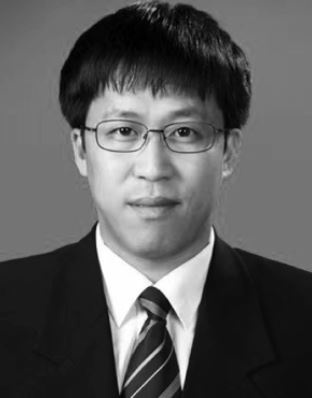}}]{Yun Lin}
(Senior Member, IEEE) received the B.S. degree in electrical engineering from Dalian Maritime
University, Dalian, China, in 2003, the M.S. degree in communication and information
system from the Harbin Institute of Technology, Harbin, China, in 2005, and the Ph.D.
degree in communication and information system from Harbin Engineering University,
Harbin, in 2010. From 2014 to 2015, he was a Research Scholar with Wright State
University, Dayton, OH, USA. He is currently a Full Professor with the College of
Information and Communication Engineering, Harbin Engineering University. He has
authored or coauthored more than 200 peer-reviewed journal and conference papers,
including papers in IEEE Transactions on Industrial Informatics, IEEE Transactions on
Communications, IEEE Internet of Things Journal, IEEE Transactions on Vehicular
Technology, IEEE Transactions on Cognitive Communications and Networking, TR, INFOCOM,
GLOBECOM, ICC, VTC, and ICNC. His current research interests include machine learning
and data analytics over wireless networks, signal processing and analysis, cognitive
radio and software-defined radio, artificial intelligence, and pattern recognition.
\end{IEEEbiography}
\begin{IEEEbiography}[{\includegraphics[width=1in,height=1.25in,clip,keepaspectratio]{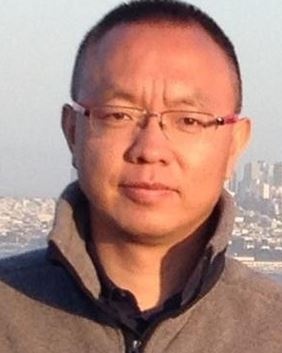}}]{Qi Xuan}
(Senior Member, IEEE) received the B.S. and Ph.D. degrees in control theory and
engineering from Zhejiang University, Hangzhou, China, in 2003 and 2008, respectively.
He was a Postdoctoral Researcher with the Department of Information Science and
Electronic Engineering, Zhejiang University from 2008 to 2010, and a Research Assistant
with the Department of Electronic Engineering, City University of Hong Kong, Hong Kong,
in 2010 and 2017, respectively. From 2012 to 2014, he was a Postdoctoral Fellow with the
Department of Computer Science, University of California at Davis, Davis, CA, USA. He is
currently a Professor with the Institute of Cyberspace Security, College of Information
Engineering, Zhejiang University of Technology, Hangzhou, and also with the PCL Research
Center of Networks and Communications, Peng Cheng Laboratory, Shenzhen, China. He is
also with Utron Technology Company Ltd., Xi’an, China, as a Hangzhou Qianjiang
Distinguished Expert. His current research interests include network science, graph data
mining, cyberspace security, machine learning, and computer vision.
\end{IEEEbiography}
\begin{IEEEbiography}[{\includegraphics[width=1in,height=1.25in,clip,keepaspectratio]{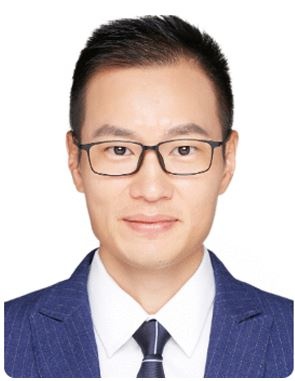}}]{Guan Gui}
    received the Dr. Eng. degree in information and communication engineering from the
    University of Electronic Science and Technology of China, Chengdu, China, in 2012.
    From 2009 to 2012, with financial support from the China Scholarship Council and the
    Global Center of Education, Tohoku University, he joined the Wireless Signal
    Processing and Network Laboratory (Prof. Adachi's laboratory), Department of
    Communications Engineering, Graduate School of Engineering, Tohoku University, as a
    research assistant and a postdoctoral research fellow. From 2012 to 2014, he was
    supported by a Japan Society for the Promotion of Science fellowship as a
    postdoctoral research fellow. From 2014 to 2015, he was an Assistant Professor with
    the Department of Electronics and Information System, Akita Prefectural University.
    Since 2015, he has been a Professor with the Nanjing University of Posts and
    Telecommunications, Nanjing, China. He is currently involved in the research of big
    data analysis, multidimensional system control, super-resolution radar imaging,
    adaptive filter, compressive sensing, sparse dictionary design, channel estimation,
    and advanced wireless techniques. He received the IEEE International Conference on
    Communications Best Paper Award in 2014 and 2017 and the IEEE Vehicular Technology
    Conference (VTC-spring) Best Student Paper Award in 2014. He was also selected as a
    Jiangsu Special Appointed Professor, as a Jiangsu High-Level Innovation and
    Entrepreneurial Talent, and for 1311 Talent Plan in 2016. He has been an Associate
    Editor of the Wiley Journal Security and Communication Networks since 2012.
\end{IEEEbiography}

\end{document}